\documentclass{article}
\usepackage{geometry}
\usepackage{pifont}
\usepackage{floatrow}
\usepackage{caption}
\usepackage{mathrsfs}
\usepackage{booktabs}
\usepackage{amsmath}
\usepackage{amsfonts,amsthm,amssymb,mathrsfs,bbding}
\usepackage{graphics,multicol}
\usepackage{graphicx}
\usepackage{color}
\usepackage{caption}
\usepackage{subcaption} 
\usepackage{indentfirst}
\usepackage{cite}
\usepackage{latexsym,bm}
\usepackage{enumerate}
\usepackage{epsfig}
\usepackage{tikz}
\usepackage{hyperref,todonotes,comment}
\usepackage{empheq,cases}
\usepackage{algorithm}
\usepackage{algorithmic}
\def\bbeta{{\bm{\theta}}}
\def\x{\bm{x}}

\def\y{\bm{y}}

\def\y{\bm{y}}

\def\btheta{\bm{\theta}}

\def\@#1{{\cal #1}}

\begin{document}

\title{Gaussian process surrogate with physical law-corrected prior for multi-coupled PDEs defined on irregular geometry}

\author{Pucheng Tang$^a$, Hongqiao Wang$^{b,d}$,  Qian Chen$^c$, Wenzhou Lin$^{c*}$, Heng Yong$^{c}$ \\
[1mm]
{\it $^a$School of Artificial Intelligence, Wuhan University, Wuhan 430072, China} \\
[1mm]
{\it $^b$School of Mathematics and Statistics, Central South University, Changsha 410083, China} \\
[1mm]
{\it $^c$Institute of Applied Physics and Computational Mathematics, Beijing 100094, China} \\
[1mm]
{\it $^d$Institute of Mathematics, Henan Academy of Sciences, Zhengzhou 450046, China} \\
}

\date{}

\maketitle {\flushleft\large\bf Abstract} \\
{Parametric partial differential equations (PDEs) serve as fundamental mathematical tools for modeling complex physical phenomena, yet repeated high-fidelity numerical simulations across parameter spaces remain computationally prohibitive. In this work, we propose a physical law–corrected prior Gaussian process (LC-prior GP) for efficient surrogate modeling of parametric PDEs.
The proposed method employs proper orthogonal decomposition (POD) to represent high-dimensional discrete solutions in a low-dimensional modal coefficient space, significantly reducing the computational cost of kernel optimization compared with standard GP approaches in full-order spaces. The governing physical laws are further incorporated to construct a law-corrected prior to overcome the limitation of existing physics-informed GP methods that rely on linear operator invariance, which enables applications to nonlinear and multi-coupled PDE systems without kernel redesign.
Furthermore, the radial basis function–finite difference (RBF-FD) method is adopted for generating training data, allowing flexible handling of irregular spatial domains. The resulting differentiation matrices are independent of solution fields, enabling efficient optimization in the physical correction stage without repeated assembly.
The proposed framework is validated through extensive numerical experiments, including nonlinear multi-parameter systems and scenarios involving multi-coupled physical variables defined on different two-dimensional irregular domains to highlight the accuracy and efficiency compared with baseline approaches.}

% \tableofcontents
\section{Introduction}
Parametric differential equations (DEs), including parametric ordinary differential equations (ODEs) and parametric partial differential equations (PDEs), are fundamental tools for modeling a wide range of scientific and engineering phenomena \cite{schiesser2019time,temam2024navier,yu2022gradient}. The associated parameters, arising from physical properties, environmental factors, or other system characteristics, play a crucial role in uncertainty quantification (UQ), sensitivity analysis, and optimization. Parametric DEs also enable the study of how input variations propagate through a system and support parameter estimation and inverse problems for inferring unknown quantities from observed data \cite{tarantola2005inverse,brivio2024ptpi}. Accurate parameter estimation and analysis typically require extensive computational sampling, often involving a large number of realizations. Although both traditional numerical methods \cite{thomas2013numerical,dhatt2012finite} and more recent machine learning approaches, such as Physics-Informed Neural Networks (PINNs) \cite{raissi2019physics}, the random feature method \cite{chen2022bridging}, and the Deep Galerkin Method \cite{sirignano2018dgm}, have achieved remarkable success in solving PDEs, their computational cost becomes prohibitive in complex practical applications, as a complete re-solution is required whenever the physical parameters change \cite{mishra2018machine}.
 
{To address these challenges, surrogate modeling techniques have emerged as an effective approach for reducing the computational cost of predictive modeling in large-scale and complex systems. These data-driven models approximate the mapping between system parameters and responses based on available data. Recent advances in machine learning have led to the widespread adoption of approaches such as deep neural networks (DNNs) \cite{raissi2019physics}, neural operators (NOs) \cite{lu2021learning}, and Gaussian process regression (GPR) \cite{williams2006gaussian} for surrogate modeling. Although these methods have demonstrated strong performance across various applications \cite{kennedy2001bayesian, chen2021improved, radaideh2020surrogate}, for parametric PDEs, they typically require large training datasets of parameter–solution pairs generated by high-fidelity solvers, which limits their efficiency and accuracy in small-data regimes. To improve model generalization, physics-informed strategies \cite{karniadakis2021physics} have been incorporated to better capture the underlying physical principles. For example, physics-informed DeepONet (PI-DeepONet) \cite{wang2021learning} introduces physics-based loss functions for operator learning, while physics-enhanced deep surrogates \cite{pestourie2023physics} leverage physical information from low-fidelity data to improve performance. These approaches offer flexibility for handling unstructured problems and have been applied to a wide range of scientific computing tasks \cite{rudy2019data,tripura2023wavelet,wang2021learning}. However, they remain computationally expensive, as evaluating PDE residuals and performing iterative training procedures are typically time-consuming. This limitation highlights the need for a more flexible framework that balances accuracy, computational efficiency, and data requirements.}

{In recent years, reduced basis (RB) methods \cite{quarteroni2015reduced}, which identify a low-dimensional subspace of the solution manifold and project the governing equations onto it, have become a popular paradigm when combined with modern machine learning frameworks \cite{lucia2004reduced,pichi2024graph}. Among these, proper orthogonal decomposition (POD)-based methods have achieved notable success \cite{nekkanti2023gappy}, as they construct optimal low-dimensional bases from solution snapshots by retaining only the most significant modes. Representative approaches include the physics-reinforced neural network (PRNN) \cite{chen2021physics}, which approximates the reduced coefficients of parametrized PDEs within a reduced-basis framework for solving PDEs, and POD-DeepONet \cite{lu2022comprehensive}, which applies POD to the training data to extract basis functions for operator learning. Building on these foundational architectures, numerous effective variants have been developed to flexibly address diverse application scenarios \cite{de2013basis,hesthaven2018non,baur2011interpolatory,song2024model}.

Although DNNs have dominated recent physics-informed research due to their compatibility with automatic differentiation, Gaussian process (GP)-based approaches have also demonstrated strong competitiveness, owing to their advantages in UQ and reduced data requirements \cite{williams2006gaussian,mora2025operator}. Remarkable works include physics-informed GPR \cite{pfortner2022physics} for generalizing linear PDE solvers and Gaussian process regression with constraints (GPRC) \cite{wang2021explicit}, which improves the prediction accuracy of derivatives by exploiting the linearity of differential operators from a Bayesian perspective. However, due to the inherent property that GP is closed only under linear operators, existing studies primarily focus on incorporating physical constraints into kernel design for linear PDEs \cite{pfortner2022physics,wang2021explicit,pang2020physics}. As a result, these approaches often struggle with nonlinear or multi-coupled PDEs, as well as with efficient kernel optimization on dense discretizations. These limitations motivate the development of a new GP framework with physical constraints to flexibly handle complex PDEs while maintaining low data requirements.}

\begin{figure}[t]
	\centering
	\includegraphics[width=1\linewidth]{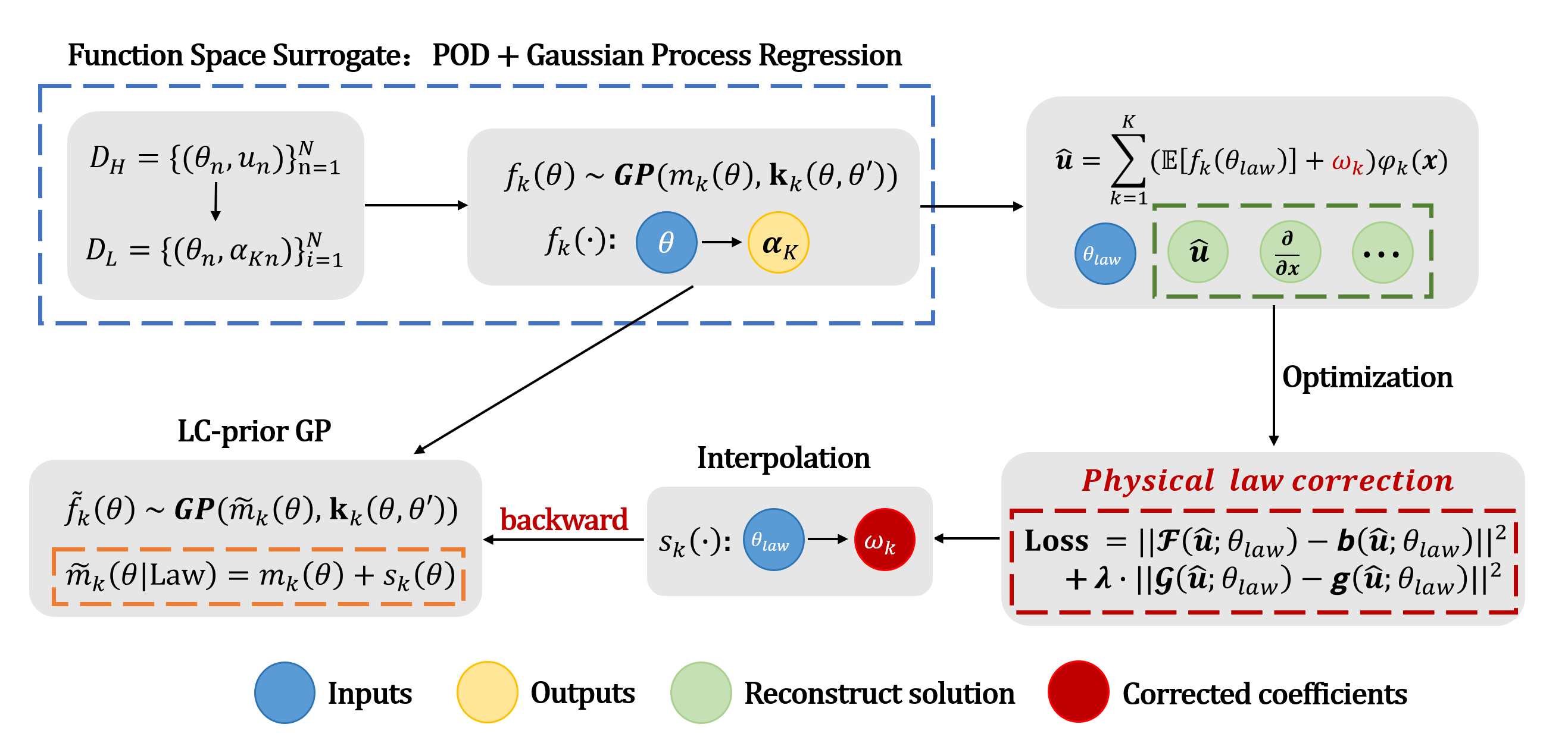}
	\caption{A Schematic of LC-prior GP for parametric differential equations.}
	\label{fig:lc-prior_gp}
\end{figure}

{Inspired by the aforementioned methods, we propose a novel \emph{physical law-corrected prior GP (LC-prior GP)} framework for constructing surrogate models for parametric PDEs in a reduced-basis representation under a small-data regime. The approach employs POD to project the infinite-dimensional solution space of the DEs onto a low-dimensional coefficient space. A GPR surrogate is then trained to map the parameters to the modal coefficients, and the predictions are further corrected using the physical laws of the PDEs, thereby learning a more consistent conditional prior that satisfies both the governing equations and the data constraints. An illustration of the LC-prior GP architecture is shown in Figure~\ref{fig:lc-prior_gp}.
In addition, to enable flexible application to irregular spatial domains, we employ the RBF-FD \cite{bayona2010rbf} method for forward simulations when generating training data. RBF-FD is a mesh-free discretization method in which the differentiation matrices, characterized by the analytical form of local stencil-based basis functions, are independent of the system parameters and can be precomputed once for a given spatial discretization \cite{shankar2017overlapped}. Although the PDE residual still needs to be evaluated during each optimization, the reuse of these matrices avoids repeated construction of differential operators with only fast matrix operations needed. This distinguishes it from DNN-based approaches relying on automatic differentiation, where derivatives must be recomputed at every iteration. Such a property provides a natural advantage in our GP framework for constructing physics-corrected states, enabling efficient and repeated evaluation of PDE derivatives.
The key contributions and advantages of this work are as follows:}

{\begin{itemize}
    \item \textbf{Reduced surrogate representation:} We propose a novel framework that combines POD with GPR to construct surrogate models in the low-dimensional modal coefficient space, significantly reducing the computational cost of kernel function optimization in high-dimensional discrete solution spaces.
    
    \item \textbf{Physical law–corrected prior GP:} By incorporating physical laws into the data-driven GP surrogate, we learn a more informative physical law-corrected prior without additional kernel optimization. This treatment overcomes the limitation of conventional physics-informed GP that relies on linear operator invariance, while preserving the low-data requirement of the GP framework.
    
    \item \textbf{Training efficiency and generalization:} By employing the RBF-FD method, the proposed framework can flexibly handle irregular domains. Since the differentiation matrices in RBF-FD are independent of the system parameters, they enable efficient optimization in the physics-based correction stage without repeated computation, which markedly distinguishes LC-prior GP from other physics-informed methods.
\end{itemize}}

The rest of this paper is structured as follows. 
Section~\ref{sec:problem} formulates the problem setup and introduces necessary preliminaries for the RBF-FD method.
Section~\ref{sec:lc-prior-gp} provides a detailed description of the implementation of the LC-prior GP, and briefly explains the parameter estimation procedure under this model.
Section~\ref{sec:numerical examples} presents numerical examples, ranging from cases involving a single quantity to multi-parameter scenarios with multi-coupled systems. 
Finally, concluding remarks are summarized in Section~\ref{sec:conclusion}.

\section{Problem setting in parametric PDEs}
\label{sec:problem}
Parametric PDEs are fundamental tools for modeling a wide range of phenomena in science and engineering. Let $\Omega$ be the computational domain with the Lipschitz continuous boundary $\partial \Omega$. The general form for differential equation with parameters can be expressed as:
\begin{equation}
	\label{eq:pde}
	\begin{cases}
		\begin{aligned}
			& \mathcal{F} \big(\bm{u}(\boldsymbol{x};\boldsymbol{\theta}),\, \nabla \bm{u},\, \nabla^{2} \bm{u},\, \dots;\, \boldsymbol{\theta}\big) = \bm{b}(\x;\btheta), \quad &\boldsymbol{x} &\in \Omega, \\ 
			& \mathcal{G} \big(\bm{u}(\boldsymbol{x};\boldsymbol{\theta}),\, \nabla \bm{u},\, \nabla^{2} \bm{u},\, \dots;\, \boldsymbol{\theta}\big) = \bm{g}(\x;\btheta), \quad &\boldsymbol{x} &\in \partial\Omega,
		\end{aligned}
	\end{cases}
\end{equation}
where $\bm{u}(\x;\btheta)$ is the solution vector and $\bm{\theta}=(\theta_1,\dots,\theta_q)^\top \in \mathbb{R}^q$ denotes the parameter vector. The operators $\mathcal{F}$ and $\mathcal{G}$ represent linear or nonlinear functions involving $\bm{u}$ and its derivatives over the domain $\Omega$ and boundary $\partial\Omega$, respectively, with $\bm{b}(\x;\btheta)$ and $\bm{g}(\x;\btheta)$ as source terms. For brevity, we write $\mathcal{F}(\bm{u}, \nabla \bm{u}, \nabla^{2} \bm{u}, \dots;, \boldsymbol{\theta})$ as $\mathcal{F}(\bm{u};\btheta)$.
We then define the operator from parameters to solutions as $\mathcal{M}: \bm{\theta} \mapsto \bm{u}(\x;\btheta).$

\subsection{RBF-FD method}
Radial basis function (RBF) methods represent a class of mesh-free techniques that can be classified into several categories. Our study focuses specifically on the local RBF-FD approach integrated with a least squares technique \cite{bayona2010rbf}. {This approach offers significant flexibility in handling complex geometries \cite{song2024model}. In this work, we focus on PDEs defined on two-dimensional spatial domains with irregular boundaries and possible interior holes, while the spatial domain remains fixed over time.}

We begin by reviewing fundamental concepts of RBF interpolation, which form the basis for the RBF-FD methodology. Consider a set of scattered nodes  $\x_i \in \mathbb{R}^d$, $i = 1, 2, \dots, n$ distributed in the neighborhood of a point $\x$, where these nodes are completely independent of any mesh or element structure. A localized RBF approximation of the function $u(\x)$ can be constructed using the radial basis functions $\phi(\|\x - \x_i\|)$,
$$
u_h(\x) = \sum_{i=1}^n c_i \phi(\|\x - \x_i\|),
$$
where $\phi(\| \cdot \|)$ is some radial function, $\| \cdot \|$ is the standard Euclidean norm, and $c_i$ are unknown coefficients. Using the interpolation condition $u_h(\x_i) = u(\x_i)$, we can obtain a linear system as:
$$
\underbrace{
	\begin{pmatrix}
		\phi(\|\x_1 - \x_1\|) & \cdots & \phi(\|\x_1 - \x_n\|) \\
		\vdots & \ddots & \vdots \\
		\phi(\|\x_n - \x_1\|) & \cdots & \phi(\|\x_n - \x_n\|)
	\end{pmatrix}
}_{A}
\underbrace{
\begin{pmatrix}
	c_1 \\
	\vdots \\
	c_n
\end{pmatrix}
}_{\mathbf{c}}
=
\underbrace{
\begin{pmatrix}
	u(\x_1) \\
	\vdots \\
	u(\x_n)
\end{pmatrix}
}_{\bm{u}}.
$$

Let $\mathbf{c} = (c_1, \dots, c_n)^\top$ and $\bm{u} = (u(\x_1), \dots, u(\x_n))^\top$. Then we have the compact form $A\mathbf{c} = \bm{u}.$

To avoid parameter tuning, we adopt piecewise smooth polyharmonic splines (PHS) for simplicity. Since PHS RBFs are conditionally positive definite, interpolation based on pure RBF alone may lead to ill-posedness and lack of convergence. To address this, the approximation is augmented with a polynomial basis of degree consistent with the order of conditional positive definiteness, together with the moment conditions imposed on the RBF coefficients $\mathbf{c}$. This ensures the non-singularity of the resulting interpolation system and guarantees a unique solution. The resulting RBF interpolation takes the form:
\begin{equation*}
	\begin{cases}
		\begin{aligned}
			&u_h(\x) = \sum_{i=1}^n c_i \phi(\|\x - \x_i\|) + \sum_{k=1}^{m} \beta_{k} p_{k}(\x), \\
			&\sum_{i=1}^n c_i p_{k}(\x_{i}) = 0,
		\end{aligned}
	\end{cases} \quad k=1,\cdots,m.
	\label{eq:RBF-interpolation}
\end{equation*}

The dimension $m$ of the polynomial space is given by $m = \binom{\text{D}_{m} + d}{d},$ where $\text{D}_{m}$ is the degree of the polynomial and $d$ is the spatial dimension of $\mathbb{R}^d$. The coefficients $c_i$ and $\beta_k$ are determined by the collocation method and the additional constraints. The numerical optimal solution can be expressed as
$$
\left(
\begin{array}{ccc|ccc}
	\phi(\|\x_1 - \x_1\|) & \cdots & \phi(\|\x_1 - \x_n\|) & p_{1}(\x_1) & \cdots & p_{m}(\x_1) \\
	\vdots & \ddots & \vdots & \vdots & \ddots & \vdots \\
	\phi(\|\x_n - \x_1\|) & \cdots & \phi(\|\x_n - \x_n\|) & p_{1}(\x_n) & \cdots & p_{m}(\x_n) \\ 
	\hline
	p_{1}(\x_1) & \cdots & p_{1}(\x_n)  & 0 & \cdots  & 0 \\
	\vdots & \ddots & \vdots & \vdots & \ddots & \vdots \\
	p_{m}(\x_1) & \cdots & p_{m}(\x_n) & 0 &  \cdots & 0
\end{array}
\right)
\left(
\begin{array}{c}
	 c_{1} \\ \vdots \\ c_n \\ \hline \beta_{1} \\ \vdots \\ \beta_{m}
\end{array}
\right)
=
\left(
\begin{array}{c}
	u(\x_1) \\ \vdots \\ u(\x_n)\\ \hline 0 \\ \vdots \\ 0
\end{array}
\right).
$$

Let $\beta=(\beta_{1},\dots,\beta_{m})^\top$, we also have the compact form as ${\hat{A}}{\hat{c}}={\hat{\bm{u}}}$:
$$
\underbrace{
\begin{pmatrix}
	{A} & {P} \\ 
	{P}^\top & \boldsymbol{0} \\
\end{pmatrix}
}_{\hat{A}}
\underbrace{
\begin{pmatrix}
	\mathbf{c} \\ \beta
\end{pmatrix}
}_{\hat{\mathbf{c}}} =
\underbrace{
\begin{pmatrix}
	\bm{u} \\ 0
\end{pmatrix}
}_{\hat{\bm{u}}}.
$$

In order to digitally discretize the problem Eq.\eqref{eq:pde}, two sets of computational points are distributed over computational domain $\Omega$:
\begin{itemize}
	\item The interpolation point set $Y = \{\y_i\}_{i=1}^N$ for generating the cardinal functions.
	\item The evaluation point set $X = \{\x_j\}_{j=1}^M$ for sampling the PDE, and $M=qN$.
\end{itemize}
\begin{figure}[t]
	\centering
	\begin{minipage}{0.35\linewidth}
		\centering
		\includegraphics[width=\linewidth]{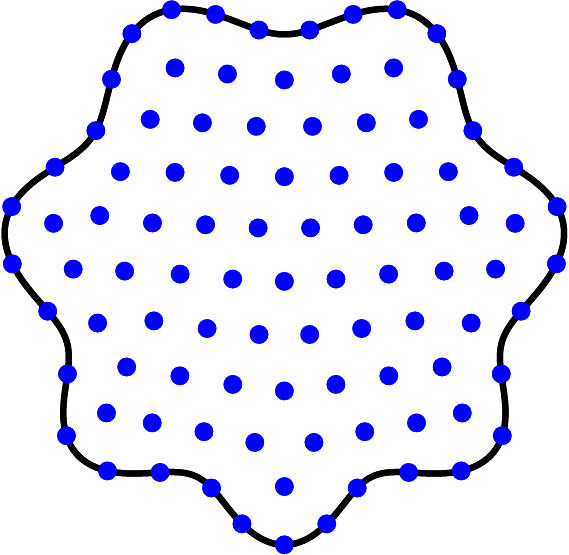}
		\subcaption{Interpolation point set $Y$}
		\label{fig:data_X}
	\end{minipage}
	\hspace{0.12\linewidth} % 添加5%行宽的间距
	\begin{minipage}{0.35\linewidth}
		\centering
		\includegraphics[width=\linewidth]{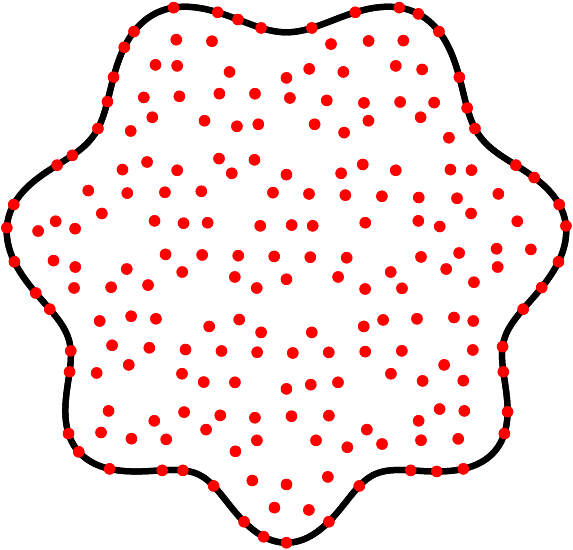}
		\subcaption{Evaluation point set $X$}
		\label{fig:data_Y}
	\end{minipage}
	\caption{The interpolation point set $Y$ and evaluation point set $X$.}
	\label{fig:data_XY}
\end{figure}

In the dataset $Y$, each data node $\y_i$ corresponds to a local support domain by the stencil $Y_s= \{\y^{s}_i \}_{i=1}^n$. 
$$
	u^{s}_h(\y) = \sum_{i=1}^n c_i \phi(\|\y - \y^{s}_i\|) + \sum_{k=1}^{m} \beta_{k} p_{k}(\y), \quad \text{if}\, \y \in X_{s}.
$$
To evaluate the RBF-FD approximation at a point $\x$, we choose the stencil associated with the point $\y_i$ that is closest to $\x$. That is,
\begin{equation}
	s(\x) = \arg\min_{i} \| \x - \y_{i} \|.
	\label{eq:stencil_distance}
\end{equation}
Any point $\x\in\Omega$ is uniquely with one stencil $Y_{s}$ by Eq.\eqref{eq:stencil_distance}. Then the local interpolation for the solution of the PDE problem evaluated at the point $\x$ with local stencil can be written as:
$$
\begin{aligned}
	u_{h}^{s}(\x) &= \sum_{i=1}^n c_i^s \phi(\|\x - \y_i^s\|) + \sum_{k=1}^m \beta_k^s p_k(\x) \\
	% &= \begin{pmatrix}
	% 	\phi(\|\x - \y_1^s\|), & \ldots, & \phi(\|\x - \y_n^s\|), & p_1(\x), & \dots, & p_m(\x)
	% \end{pmatrix}
	% \begin{pmatrix}
	% 	\boldsymbol{c}^{(s)} \\
	% 	\boldsymbol{\beta}^{(s)}
	% \end{pmatrix}  \\
	&= \begin{pmatrix}
		\phi(\|\x - \y_1^s\|), & \dots, & \phi(\|\x - \y_n^s\|), & p_1(\x), & \dots, & p_m(\x)
	\end{pmatrix}
	\begin{pmatrix}
		A^{(s)} & P^{(s)} \\
		(P^{(s)})^\top & \boldsymbol{0}
	\end{pmatrix}^{-1}
	\begin{pmatrix}
		\boldsymbol{u}^{(s)}_{h} \\
		\boldsymbol{0}
	\end{pmatrix} \\
	&= \begin{pmatrix}
		\Phi_1^s(\x), & \Phi_2^s(\x), & \ldots, & \Phi_n^s(\x), & \delta_1^s(\x), & \delta_2^s(\x), & \dots, & \delta_m^s(\x)
	\end{pmatrix}
	\begin{pmatrix}
		\boldsymbol{u}^{(s)}_{h} \\
		\boldsymbol{0}
	\end{pmatrix} \\
	&= \sum_{i=1}^{n} \Phi_{i}^{s}(\x) u_{h}(\y_{i}^{s}), \qquad \text{where} \,\, \y_{i}^{s} \in Y_{s},
\end{aligned}
$$
where the function $\Phi_{i}^{s}(\cdot)$ and $\delta_{k}^{s}(\cdot)$ can by written as:
$$
\begin{aligned}
	\Phi_{i}^{s}(\x) = \phi(\|\x - \y_i^s\|) 
	\begin{pmatrix}
		A^{(s)} & P^{(s)} \\
		(P^{(s)})^\top & \boldsymbol{0}
	\end{pmatrix}^{-1}, &\quad i=1,\cdots,n, \\
	\delta_{k}^{s}(\x) = p_{k}(\x)
	\begin{pmatrix}
		A^{(s)} & P^{(s)} \\
		(P^{(s)})^\top & \boldsymbol{0}
	\end{pmatrix}^{-1}, &\quad k=1,\cdots,m.
\end{aligned}
$$
For ease of analysis, we define the global function of the whole computational domain:
$$
	\Phi_{i}(\x) = 
	\begin{cases}
		\Phi_{i}^{s}(\x), \quad &\y_{i} \in Y_s, \\
		0, & \y_{i} \notin Y_s.
	\end{cases}
$$
For any point from evaluation point set $X$, we can use the global basis function $\Phi_{i}(\cdot)$ and Eq.\eqref{eq:stencil_distance} to construct the interpolation function by $u_{h}(\x) = \sum_{i=1}^{N} \Phi_{i}(\x) u_{h}(\y_{i})$ and construct a sparse global linear system:
$$
	\underbrace{
	\begin{pmatrix}
		\Phi_{1}(\x_1) & \cdots & \Phi_{N}(\x_1) \\
		\vdots & \ddots & \vdots \\
		\Phi_{1}(\x_M) & \cdots & \Phi_{N}(\x_M)
	\end{pmatrix} }_{E_{h}(X,Y)}
	\underbrace{
	\begin{pmatrix}
		u_{h}(\y_{1}) \\ \vdots \\ u_{h}(\y_{N})
	\end{pmatrix} }_{u_{h}(Y)} =
	\underbrace{
	\begin{pmatrix}
		u_{h}(\x_{1}) \\ \vdots \\ u_{h}(\x_{M})
	\end{pmatrix} }_{u_{h}(X)}.
$$
And we have the compact form: $u_{h}(X)=E_{h}(X,Y) u_{h}(Y)$. The expression for the action of a differential operator $\mathcal{L}$ on the RBF approximation $u_{h}(\x)$ as follow:
$$
	\mathcal{L}	u_{h}(\x) = \sum_{i=1}^{N} \mathcal{L} \Phi_{i}(\x) u_{h}(\y_{i}),
$$
as well as the global sampling set $X$: 
$$\mathcal{L} u_{h}(X) = D^{\mathcal{L}}_{h}(X,Y) u_{h}(Y).$$

The evaluation matrix $E_{h}(X,Y)$ and differentiation matrix $D^{\mathcal{L}}_{h}(X,Y)$ are both $M\times N$ sparse matrices. Using the RBF for the spatial discretization of the Eq.\eqref{eq:pde} system, we can obtain:
$$
	\begin{cases}
		\begin{aligned}
			\mathcal{F}(E_{h}(X,Y) u_{h}(Y), D^{\nabla}_{h}(X,Y) u_{h}(Y)
			, D^{\nabla^{2}}_{h}(X,Y) u_{h}(Y), \cdots;\btheta) &= \bm{b}(X;\btheta), \quad X\in\Omega, \\
			\mathcal{G}(E_{h}(X,Y) u_{h}(Y), D^{\nabla}_{h}(X,Y) u_{h}(Y)
			, D^{\nabla^{2}}_{h}(X,Y) u_{h}(Y), \cdots;\btheta) &= \bm{g}(X;\btheta), \quad X\in\partial \Omega.
		\end{aligned}
	\end{cases}
$$
In order to solve the solution $u_{h}(Y)$, we can construct a linear system and use least squares method. For the sake of intuitive expression, the boundary points are not distinguished here:
$$
\left(
	\begin{array}{ccc} 
		\mathcal{F}(\Phi_{1}(\x_1), \nabla \Phi_{1}(\x_1), \cdots;\btheta) & \cdots &  
		\mathcal{F}(\Phi_{N}(\x_1), \nabla \Phi_{N}(\x_1), \cdots;\btheta) \\
		\mathcal{F}(\Phi_{1}(\x_2), \nabla \Phi_{1}(\x_2), \cdots;\btheta) & \cdots &  
		\mathcal{F}(\Phi_{N}(\x_2), \nabla \Phi_{N}(\x_2), \cdots;\btheta) \\
		\vdots &  \ddots & \vdots \\
		\mathcal{F}(\Phi_{1}(\x_M), \nabla \Phi_{1}(\x_M), \cdots;\btheta) & \cdots &  
		\mathcal{F}(\Phi_{N}(\x_M), \nabla \Phi_{N}(\x_M), \cdots;\btheta) \\
	\end{array} 
\right)
\left(
	\begin{array}{c}
		u_{h}(\y_{1}) \\ u_{h}(\y_{2}) \\ \vdots \\ u_{h}(\y_{N})
	\end{array}
\right) = 
\left(
	\begin{array}{c}
		{b}(\x_{1};\btheta) \\ {b}(\x_{2};\btheta) \\ \vdots \\	{b}(\x_{M};\btheta)
	\end{array}
\right) .
$$

After calculating the interpolation point domain $u_{h}(Y)$, the evaluation point domain can be naturally calculated by evaluation matrix $E_{h}(X,Y)$: $\bm{u}(X)=E_{h}(X,Y)u_{h}(Y)$.

The RBF-FD method is equally applicable for solving time-dependent dynamic PDEs. To illustrate this capability, we examine the following generalized temporal equation form:
\begin{equation}
	\label{eq:pde_time}
	\begin{cases}
		\begin{aligned}
			& \frac{\partial}{\partial t} \bm{u}(\x,t;\btheta)  = \mathcal{F} \big(\bm{u},\, \nabla \bm{u},\, \nabla^{2} \bm{u},\, \cdots;\, \boldsymbol{\theta}\big) , \quad  &\boldsymbol{x} \in \Omega, t\in  [0,T], \\ 
			&  \bm{u}(\x,0;\btheta) = \bm{u}_{0}(\x;\btheta)  \quad  &\boldsymbol{x} \in \Omega .
		\end{aligned}
	\end{cases}
\end{equation}

The construction procedures for RBF, evaluation and differentiation matrices remain rigorously consistent with the aforementioned methodology. The essential distinction manifests primarily in the matrix assembly strategy of the linear solving system:
$$
	\frac{\partial}{\partial t}E_{h}(X,Y) {u}_{h}(Y,t_{n+1};\btheta) = \mathcal{F}(E_{h}(X,Y) u_{h}(Y,t_{n}), D^{\nabla}_{h}(X,Y) u_{h}(Y,t_{n})
	, D^{\nabla^{2}}_{h}(X,Y) u_{h}(Y,t_{n}), \cdots;\btheta).
$$
Take the Forward-Euler method with $\Delta t$ for temporal discretization as an example, the Eq.\eqref{eq:pde_time} reads
$$
		\begin{pmatrix}
			\Phi_{1}(\x_1) & \cdots & \Phi_{N}(\x_1) \\
			\Phi_{1}(\x_2) & \cdots & \Phi_{N}(\x_2) \\
			\vdots & \ddots & \vdots \\
			\Phi_{1}(\x_M) & \cdots & \Phi_{N}(\x_M)
	\end{pmatrix}
		\begin{pmatrix}
			(u_{h}(\y_{1},t_{n+1}) - u_{h}(\y_{1},t_{n}))/\Delta t \\
			(u_{h}(\y_{2},t_{n+1}) - u_{h}(\y_{2},t_{n}))/\Delta t \\
			\vdots \\
			(u_{h}(\y_{N},t_{n+1}) - u_{h}(\y_{N},t_{n}))/\Delta t \\
	\end{pmatrix} =
		\begin{pmatrix}
			\mathcal{F}(u_{h}(\x_{1},t_{n}), \nabla u_{h}(\x_{1},t_{n}), \cdots;\btheta) \\
			\mathcal{F}(u_{h}(\x_{2},t_{n}), \nabla u_{h}(\x_{2},t_{n}), \cdots;\btheta) \\
			\vdots \\
			\mathcal{F}(u_{h}(\x_{M},t_{n}), \nabla u_{h}(\x_{M},t_{n}), \cdots;\btheta) 
	\end{pmatrix}.
$$

Through temporal discretization and the construction of the aforementioned linear system, the RBF-FD method can effectively transform complex dynamic PDE into a system of ODE in the temporal domain, which is then solved iteratively. By calculating the interpolation point domain $u_h(Y,t_{n+1};\btheta)$, the evaluation point domain at the same time $t_{n+1}$ can be calculated: $\bm{u}(X,t_{n+1};\btheta)=E_{h}(X,Y)u_h(Y,t_{n+1};\btheta)$.

In various practical applications, once the parameters change, the solution must be recomputed via $\mathcal{M}_{\text{RBF}}$, which becomes impractical when direct numerical solutions of PDEs using $\mathcal{M}_{\text{RBF-FD}}$ are prohibitively expensive. This limitation motivates the construction of a surrogate map 
$\mathcal{M}_{\text{sur}}: \btheta \mapsto \bm{u}(\x;\btheta),$ 
which efficiently approximates the underlying numerical solution operator using a limited set of training data 
\( \mathcal{D} = \left\{ \left(\btheta_{n}, \, \bm{u}(\x;\btheta_{n}) \right) \right\}_{n=1}^{N} \).

\section{Law-corrected surrogate modeling with Gaussian process}
\label{sec:lc-prior-gp}
{In this section, we present the details of the LC-prior GP for surrogate modeling of parametric PDEs. The target problem is Eq.~\eqref{eq:pde}, where we aim to learn the mapping $\mathcal{M}_{\text{sur}}: \btheta \mapsto u(\x;\btheta)$ by an efficient and accurate surrogate. We first introduce the reduced representation of parametric PDEs using POD, and then describe in detail how to learn the physics-corrected prior function based on a data-driven GP surrogate, where nearly analytical accuracy in differential operations is achieved through the differentiation matrices in the RBF-FD method. Finally, based on the proposed method, we briefly present parameter estimation within a Bayesian framework.}

\subsection{Parametric representation of the solution for PDEs}

Basis function expansion is a widely used technique for representing complex functions \cite{schaback2024using}. By linearly combining basis functions, various functions can be approximated or reconstructed.

We construct an approximate PDE solution, denoted as \( \hat{\bm{u}}(\x; \boldsymbol{\theta}) \), using \( K \) orthogonal basis functions:
$$\bm{u}(\x; \boldsymbol{\theta}) \approx \hat{\bm{u}}(\x; \boldsymbol{\theta}) = \sum_{k=1}^{K} \alpha_{k}(\boldsymbol{\theta}) \, \phi_{k}(\x),$$
where $\phi_k(\x)$ is the basis function and \( \alpha_{k}(\btheta) \) is the coefficient corresponding to the orthogonal basis. Once the type of basis function and the truncation value \( K \) are determined, the function \( \bm{u}(\x; \boldsymbol{\theta}) \) is completely represented by $K$ coefficients.
Different types of basis functions are suitable for different applications, and choosing the appropriate basis can significantly enhance computational efficiency and accuracy. 

\subsubsection{Proper orthogonal decomposition}

To achieve an efficient representation of the target functions, conventional basis functions (e.g., Fourier bases and Hermite bases) often fail to adequately capture the full features of PDE solutions when only a limited number of basis functions is used. To address this fundamental challenge, we employ Proper Orthogonal Decomposition (POD) \cite{berkooz1993proper}, which distinguishes itself from traditional basis function approaches by eliminating the need for prior assumptions about the form of the basis functions. {Instead, POD derives optimal basis functions directly from system data through the singular value decomposition of the snapshot matrix constructed from training data \cite{nguyen2023proper}. The resulting basis functions are mutually orthogonal, and the expansion coefficients are decoupled through orthogonal projection.}

Suppose a set of high-dimensional training data $\mathcal{D}_{\text{High}}=\{(\btheta_{n}, \bm{u}(\x;\btheta_{n}))\}^{N}_{n=1}$ contains parameters $\btheta_{n} \in \mathbb{R}^q$, and $\x=(x_1, \cdots, x_{D})^\top \in \mathbb{R}^D$ represents the discretized spatial domain. By discretizing the solutions at $D$ spatial points for each parameter $\btheta_n$, we construct an $N \times D$ snapshot matrix $\bm{U}$ as follows:
$$\bm{U} = 
\begin{pmatrix}
	\bm{u}(\x; \btheta_{1}) \\
	\vdots \\
	\bm{u}(\x; \btheta_{N})
\end{pmatrix} = 
\begin{pmatrix}
	u(x_1; \btheta_{1}) & \cdots & u(x_D; \btheta_{1}) \\
	\vdots & \ddots & \vdots \\
	u(x_1; \btheta_{N}) & \cdots & u(x_D; \btheta_{N})
\end{pmatrix}.
$$
where each row of $U$ represents the discrete solution \(\bm{u}(\x;\btheta_{n})\) discretized across the spatial domain for a given parameter $\btheta_{n}$ and we compute the covariance matrix of snapshot matrix as:
$$\bm{C} = \frac{1}{N -1} \bm{U}^\top \bm{U}.$$

{We perform eigenvalue decomposition of $C$ as $C\phi_{k}=\lambda_{k} \phi_{k}$, where $\{\lambda_{k}\}_{k=1}^{D}$ are the eigenvalues in descending order and $\{\phi_{k}\}_{k=1}^{D}$ are the corresponding eigenvectors.} The eigenvector $\phi_k$ also corresponds to the $k$-th POD mode, while the associated eigenvalue $\lambda_k$ quantifies its energy contribution to the snapshot matrix. Larger eigenvalues indicate more significant modes. Thus, the leading $K$ eigenvectors $\bm{\phi}=({\phi}_{1},\dots,{\phi}_K)^\top$ are selected as basis functions to approximate the solution.
$$
\underbrace{
\begin{pmatrix}
	\alpha_{1}(\btheta_{1}) & \cdots & \alpha_{K}(\btheta_{1} ) \\
	\alpha_{1}(\btheta_{2}) & \cdots & \alpha_{K}(\btheta_{2} ) \\
	\vdots & \ddots & \vdots \\
	\alpha_{1}(\btheta_{N}) & \cdots & \alpha_{K}(\btheta_{N} ) 
\end{pmatrix}}_{N\times K}
\underbrace{
\begin{pmatrix}
	\phi_{1}(x_{1}) & \cdots & \phi_{1}(x_{D}) \\
	\phi_{2}(x_{1}) & \cdots & \phi_{2}(x_{D}) \\
	\vdots & \ddots & \vdots \\
	\phi_{K}(x_{1}) & \cdots & \phi_{K}(x_{D}) 
\end{pmatrix}}_{K\times D}
= 
\underbrace{
\begin{pmatrix}
	{u}(x_1; \btheta_{1}), & \dots, & {u}(x_D; \btheta_{1}) \\
	{u}(x_1; \btheta_{2}), & \dots, & {u}(x_D; \btheta_{2}) \\
	\vdots & \ddots & \vdots \\
	{u}(x_1; \btheta_{N}), & \dots, & {u}(x_D; \btheta_{N}) 
\end{pmatrix}}_{N\times D}
$$
where $\bm{\alpha}(\btheta_n)=(\alpha_1(\btheta_n),\dots,\alpha_K(\btheta_n))^\top$ denotes the coefficient vector corresponding to the $n$-th row, which can be computed via the least squares method. The original matrix $\bm{U}$ can then be approximated by a linear combination of the POD basis functions (eigenvectors) and the corresponding coefficients as follows:
$$
\bm{U} = \begin{pmatrix}
u(\x; \btheta_{1}) \\
\vdots \\
u(\x; \btheta_{N})
\end{pmatrix} \approx 
\begin{pmatrix}
	\sum^{K}_{k=1}\alpha_{k}(\btheta_{1}) \phi_{k}(x_{1}) & \cdots & \sum^{K}_{k=1}\alpha_{k}(\btheta_{1}) \phi_{k}(x_{D}) \\
	\vdots & \ddots & \vdots \\
	\sum^{K}_{k=1}\alpha_{k}(\btheta_{N}) \phi_{k}(x_{1}) & \cdots & \sum^{K}_{k=1}\alpha_{k}(\btheta_{N}) \phi_{k}(x_{D})
\end{pmatrix}.
$$
{To determine the optimal number of POD modes $K$, we define the cumulative energy capture ratio $\eta(K)$ as
\begin{equation} 
    \label{eq:pod_energy}
    \eta(K) = \frac{\sum_{k=1}^K \lambda_k}{\sum_{k=1}^D \lambda_k},
\end{equation}
where $\eta(K)$ represents the fraction of total energy captured by the first $K$ modes. The expansion is truncated at the smallest $K$ such that $ \eta(K) > 99.99\%$. To assess the reconstruction accuracy, we introduce the relative $L^1$ error
\begin{equation}
\label{eq:norm1_error}
    \text{Error}=\frac{1}{N}\sum_{n=1}^{N} \frac{\left\| \hat{\bm{u}}(\x,\btheta_n) - {\bm{u}}(\x,\btheta_n)  \right\|_1}{\left\|{\bm{u}}(\x,\btheta_n)  \right\|_1},
\end{equation}
where $\|\cdot\|_1$ denotes the discrete $L^{1}$ norm. The $\bm{u}(\x;\btheta_n)$ and $\hat{\bm{u}}(\x;\btheta_n)=\sum^{K}_{k=1}\alpha_{k}(\btheta_{n}) \phi_{k}(\x)$ denote the original data and the corresponding reconstruction obtained by POD, respectively. Under this strategy, when the truncated energy satisfies $ \eta(K) > 99.99\%$, the relative $L^{1}$ error between the approximated and true solutions is observed to be significantly smaller than this threshold. This level of accuracy is sufficient for constructing the surrogate model, thereby ensuring high-fidelity reconstruction with a minimal number of modes.
}

Through POD, we map the $D$-dimensional discrete solution space to a $K$-dimensional coefficient space:
$$
\text{POD}:\,\bm{u}(\x;\btheta_{n}) \in \mathbb{R}^D \mapsto \bm{\alpha}(\btheta_{n}) \in \mathbb{R}^K, \quad n=1,\dots,N,\quad K \ll D.
$$
By fixing the POD modes $\bm{\phi}$ as basis functions, we can reconstruct the solution corresponding to any parameter $\btheta$ by predicting the associated coefficients $\bm{\alpha}(\btheta)$. Therefore, the task is to learn a surrogate model that represents the mapping from parameters to POD coefficients using the low-dimensional training data $\mathcal{D}_{\text{Low}} = \left\{ \left(\btheta_{n}, \bm{\alpha}_{n} \right) \right\}_{n=1}^{N}$.

\subsection{Gaussian process regression surrogate model}
 {Gaussian process regression (GPR) is nonparametric framework for surrogate modeling\cite{williams2006gaussian}. Our objective is to learn a mapping $f: \btheta \in \mathbb{R}^{q} \mapsto \bm{\alpha}(\btheta) \in \mathbb{R}^{K}$, by the reduced-order training dataset
$$
\mathcal{D}_{\text{Low}} = \left\{ \left(\btheta_{n}, \bm{\alpha}_{n} \right) \right\}_{n=1}^{N}, \quad \text{with} \quad \bm{\alpha}_{n}=(\alpha_{n1},\dots,\alpha_{nK})^\top \in \mathbb{R}^K,
$$

Since POD provides an orthogonal basis, each modal coefficient corresponds to the contribution along a distinct basis direction in the reduced-order representation. Modeling them independently avoids introducing artificial correlations through the kernel and yields a more efficient and scalable surrogate. Therefore, we employ $K$ independent GPR models, each associated with one modal coefficient. For the $k$-th mode, we define $f_{k}: \btheta \in \mathbb{R}^{q} \mapsto \alpha_{k}(\btheta) \in \mathbb{R}$, and denote the coefficients of the $k$-th mode across all samples by $\bm{\alpha}_{k} = (\alpha_{1k}, \dots, \alpha_{Nk})^\top \in \mathbb{R}^N$. Each model follows a Gaussian process:
$$
f_{k}(\btheta) \sim \mathcal{GP}\left(m_{k}(\btheta), \mathbf{k}_{k}(\btheta, \btheta') \right), \quad k=1,\dots,K.
$$}
where $m_{k}(\btheta)$ denotes the mean function, typically assumed to be zero, and $\mathbf{k}_{k}(\btheta, \btheta')$ is the covariance kernel. A commonly used choice is the RBF kernel read as $\mathbf{k}_{\text{RBF}} = \gamma^2 \exp\left(-\frac{\|\btheta_{i} - \btheta_{j}\|^2}{2\ell^2}\right).$

According to the definition of GP, the finite projection of $f_{k}(\cdot)$ onto the training inputs $\bm{\theta}$, namely $\bm{f}_{k} = (f_{k}(\btheta_1), \dots, f_{k}(\btheta_N))^\top$, follows a multivariate Gaussian distribution, 
$$
p(\bm{f}_{k}|\btheta) = \mathcal{N}(\bm{f}_{k}|\bm{0}, \text{\bf{K}}_{k}),
$$
where $\text{\bf{K}}_{k}$ represents the kernel matrix evaluated at $\bm{\theta}$ and each element is defined as $[\text{\bf{K}}_{k}]_{i,j} = \text{\bf{k}}_{k}(\bm{\theta}_i, \bm{\theta}_j)$. Let $\bm{\zeta}_{k}=(\gamma_{k}, \ell_{k})$ denote the set of all hyper-parameters associated with $\text{\bf{K}}_{k}$. To learn the model, we maximize the log-likelihood to estimate the kernel parameters $\bm{\zeta}_{k}$ for each $\bm{\alpha}_{k}$:
$$
\log p(\bm{\alpha}_{k}|\btheta, \bm{\zeta}_{k}) = -\frac{1}{2} \bm{\alpha}_{k}^\top \text{\bf{K}}_{k}(\btheta,\btheta')^{-1} \bm{\alpha}_{k} - \frac{1}{2} \log \det \text{\bf{K}}_{k}(\btheta,\btheta') - \frac{N}{2} \log 2\pi.
$$

According to the GP prior, given a new input $\btheta^*$, the posterior (or predictive) distribution of the output $f_k(\btheta^*)$ is a conditional Gaussian distribution:
$$p\big(f_{k}(\btheta^*)|\btheta^*, \mathcal{D}_{\text{Low}} \big) = \mathcal{N}\big(f_{k}(\btheta^*)|\, \mu_{k}(\btheta^*), \sigma^{2}_{k}(\btheta^*)\big),$$
where the posterior mean and variance are given by:
\begin{equation} \label{eq:gp_poster}
	\begin{aligned} 
		\mu_{k}(\btheta^*)  &= \mathbb{E}[f_{k}(\btheta^*)|\btheta^*, \mathcal{D}_{\text{Low}}]  
		= m_{k}(\btheta^*) + \text{\bf{k}}_{k*}^\top \text{\bf{K}}^{-1}_{k} \big( \bm{\alpha}_{k}- \mathbf{m}_{k}  \big) = \text{\bf{k}}_{k*}^\top \text{\bf{K}}^{-1}_{k} \bm{\alpha}_{k},      \\
		\sigma^{2}_{k}(\btheta^*) &= \text{Var}[f_{k}(\btheta^*)|\btheta^*, \mathcal{D}_{\text{Low}}]  
		= \text{\bf{k}}_{k}(\btheta^*, \btheta^*) - \text{\bf{k}}_{k*}^{\top} \text{\bf{K}}_{k}^{-1} \text{\bf{k}}_{k*},
	\end{aligned}
\end{equation}
where $\mathbf{m}_{k}=(m_k(\btheta_1),\dots,m_k(\btheta_N))^\top$ denotes the prior mean vector evaluated at the training inputs, and $ \text{\bf{k}}_{k*} = (\text{\bf{k}}_{k}(\btheta^*, \btheta_1), \dots, \text{\bf{k}}_{k}(\btheta^*, \btheta_N))^\top$ represents the kernel evaluations between $\btheta^*$ and the input.

By iterating this process, we independently learn the surrogate model \( f_{k} \) for each modal coefficient \( {\alpha}_{k}(\btheta) \). The solution \( \bm{u}(\x; \btheta^*) \) is then predicted as a linear combination of the \( K \) fixed basis functions with the corresponding predicted coefficients given by the posterior mean
\begin{equation}
	\hat{\bm{u}}(\x;\btheta^*) = \mathcal{M}_{\text{GP}}(\x;\btheta^*) = \sum_{k=1}^{K} \mu_{k}(\btheta^*) \phi_{k}(\x).
	\label{eq:M_gp}
\end{equation}

\subsection{Prior correction with physical laws}

 {Based on physics-informed methods, some GP approaches that incorporate physical constraints have achieved promising results for linear PDEs in recent years \cite{pfortner2022physics,wang2021explicit}. However, such methods typically exploit the linearity of GP by imposing linear transformations on the kernel function to encode physical constraints, which makes them difficult to extend to nonlinear or multi-coupled PDEs and significantly increases the burden of hyperparameter optimization. To address this, we propose embedding physical laws as prior knowledge directly into the learned GP surrogate, enabling more flexible modeling of various parametric PDEs.}

For the standard GP surrogates, the prior mean function specified during training is still used at the prediction stage. Thus, a straightforward idea is to learn a more reasonable prior mean function $\tilde{m}_k(\btheta|\text{Law})$ under the constraints of physical laws and the learned model. Therefore, a correction function ${\omega_{k}}(\btheta|\text{Law})$ is introduced to adjust the original prior mean function, and we can define a novel physical law-corrected prior (LC-prior) as
$$\tilde{m}_k(\btheta|\text{Law}) = m_{k}(\btheta) + \omega_k(\btheta|\text{Law}),$$ 
where $m_{k}(\btheta)$ is prior mean function of $f_{k}(\cdot)$ that is general supposed to constant $0$ and the $\omega_k(\btheta|\text{Law})$ is correction function that needs to be learned by physical law.
\begin{figure}[t]
	\centering
	\includegraphics[width=0.8\linewidth]{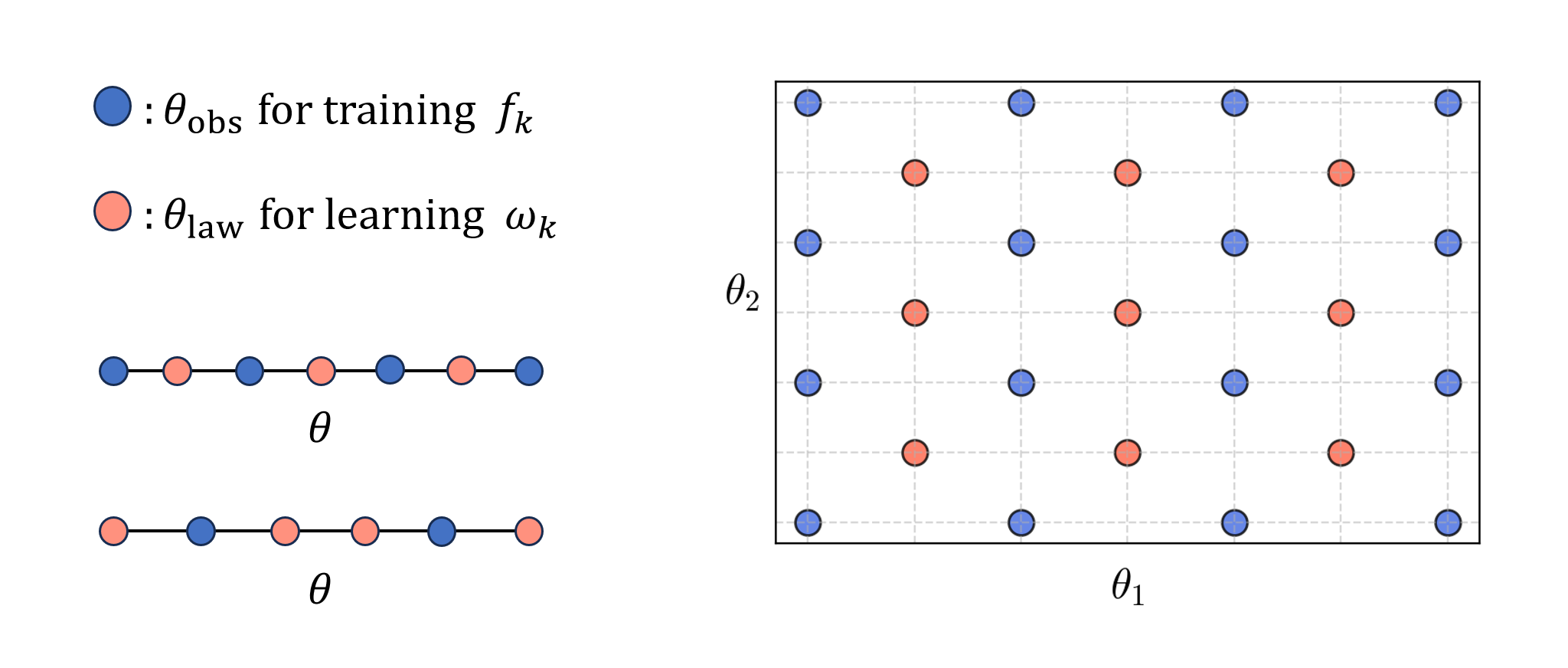}
	\caption{ {A basic strategy for selecting $\btheta_{\text{law}}$ in 1D (left) and 2D (right) parameter spaces. Blue points indicate the observed parameters $\btheta_{\text{obs}}$ used to train $f_k(\cdot)$, while orange points denote the additional samples $\btheta_{\text{law}}$ for learning the physical law correction function $\omega_k(\cdot)$.}}
	\label{fig:points}
\end{figure}

To distinguish between training and unseen data, let $\btheta_{\text{obs}} \subset \{\btheta\mid(\btheta,\bm{\alpha}) \in \mathcal{D}_{\text{Low}}\}$ denote the set of training inputs used to learn the data-driven GP model. Keeping the zero prior assumption unchanged for the training data, the physical law-corrected prior function can be further decomposed into two parts as
\begin{equation} \label{eq:lc_prior}
\tilde{m}_k(\btheta|\text{Law}) =
\begin{cases}
	0, & \btheta \in \btheta_{\text{obs}}, \\
	\omega_{k}(\btheta|\text{Law}), & \btheta \in \Theta \setminus \btheta_{\text{obs}}.
\end{cases}
\end{equation}

By introducing the LC-prior, it is possible to construct a novel LC-prior GP surrogate $\tilde{f}_{k}(\cdot)$ that simultaneously leverages data-driven and physical constraints reads
$$
	\tilde{f}_{k}(\btheta) \sim \mathcal{GP}\left(\tilde{m}_{k}(\btheta|\text{Law}), \mathbf{k}_{k}(\btheta, \btheta') \right).
$$
For given any new parameter $\btheta^*$, we can subsequently derive the conditional posterior mean by
\begin{equation} \label{eq:lc_posterior}
\begin{aligned}
	   \tilde{\mu}_{k}(\btheta^*)  &= \mathbb{E}[f_{k}(\btheta^*)|\btheta^*, \mathcal{D}_{\text{Low}}, \text{Law}] 
	= \tilde{m}_{k}(\btheta^*|\text{Law}) + \text{\bf{k}}_{k*}^\top \text{\bf{K}}^{-1}_{k} \big( \bm{\alpha}_{k}- \mathbf{m}_{k}  \big)  \\ &= \omega_{k}(\btheta^*|\text{Law}) + \text{\bf{k}}_{k*}^\top \text{\bf{K}}^{-1}_{k} \bm{\alpha}_{k}
    = \omega_{k}(\btheta^*|\text{Law}) + {\mu}_{k}(\btheta^*)
\end{aligned}
\end{equation}
Similar to the Eq.\eqref{eq:M_gp}, we can likewise approximate the function $\bm{u}(\x;\btheta^{*})$ with physical law 
\begin{equation} \label{eq:M_lc_old}
	\begin{aligned}
		\hat{\bm{u}}(\x;\btheta^*) &= \mathcal{M}_{\text{LC}}(\x;\btheta^*) 
		= \sum_{k=1}^{K} \tilde{\mu}_{k}(\btheta^*) \phi_{k}(\x)  
		% = \sum_{k=1}^{K}  \big(\mu_{k}(\btheta^*)+\omega_{k}(\btheta^*) \big) \phi_{k}(\x) 
		= \sum_{k=1}^{K} \omega_{k}(\btheta^*|\text{Law}) \phi_{k}(\x) + \mathcal{M}_{\text{GP}}(\x;\btheta^*).
	\end{aligned}
\end{equation}

To learn the mapping from parameters to correction coefficients $\omega_k$, we extract an additional subset of $N_{\text{law}}$ physics-corrected points, $\btheta_{\text{law}} \subset \Theta \setminus \btheta_{\text{obs}}$, which is employed to train the correction terms.  {A common strategy for selecting the physics-corrected points $\btheta_{\text{law}}$ is to uniformly sample from regions where $\btheta_{\text{obs}}$ is absent, in order to enrich the information in unexplored areas of the parameter space $\Theta$. In our experiments, we observe that using up to approximately twice the number of training data per parameter dimension is sufficient to achieve a good balance between accuracy and computational efficiency, while denser sampling tends to introduce unnecessary computational overhead. For limited data or extrapolation scenarios, a moderately increased number of $\btheta_{\text{law}}$ is beneficial. Figure~\ref{fig:points} illustrates examples of selecting $\btheta_{\text{law}}$ for cases with $\btheta \in \mathbb{R}$ and $\btheta \in \mathbb{R}^2$.} 

The physical law loss function \cite{raissi2019physics} is given by the residual of the governing PDE in Eq.~\eqref{eq:pde} as
\begin{equation}
	\text{Loss} = || \mathcal{F}(\mathcal{M}_{\text{LC}}(\x;\btheta_{\text{law}})) - \bm{b}(\x;\btheta_{\text{law}}) ||^{2}_{2} + \bm{\lambda} \cdot || \mathcal{G}(\mathcal{M}_{\text{LC}}(\x;\btheta_{\text{law}})) - \bm{g}(\x;\btheta_{\text{law}}) ||^{2}_{2},
	\label{eq:Loss_function}
\end{equation}
where $\|\cdot\|_2$ denotes the $L^2$ norm and penalty weights will be set as $\bm{\lambda}=100$ in the following experiments. 

 {Benefiting from the UQ capability of GP, predictions naturally provide both the posterior mean and variance. We can perform a bounded optimization within the range $[-z \cdot \sigma_{k}(\btheta_{\text{law}}),\, z \cdot \sigma_{k}(\btheta_{\text{law}})]$ to optimize $\omega_k$, where $\sigma_{k}(\btheta_{\text{law}})$ denotes the standard deviation of each prediction. Since $[\mu_{k}-2\sigma_{k},\, \mu_{k}+2\sigma_{k}]$ corresponds to a $95\%$ confidence interval of the posterior distribution, it is highly likely to contain the optimal correction coefficient. Therefore, we set $z=2$ in the following experiments and the optimizer is chosen to be L-BFGS-B \cite{zhu1997algorithm}. }

 {In the optimization of the loss function \eqref{eq:Loss_function}, whether using numerical methods or automatic differentiation based on DNNs, the differential operator $\mathcal{F}$ must be recomputed at every iteration, which incurs high computational cost. However, this issue can be effectively addressed by the RBF-FD method: the differentiation matrices $D^{\nabla}_{h}(X,Y)$, $D^{\nabla^{2}}_{h}(X,Y)$, etc., in sparse matrix form, depend only on the scattered node locations and are independent of the system parameters. In particular, although the operator $\mathcal{F}$ is still evaluated at each iteration, all derivative terms of any order can be computed via the precomputed differentiation matrices using matrix operations. This property enables the LC-prior GP to avoid expensive recomputation of derivatives, allowing the Eq.~\eqref{eq:Loss_function} to be computed more efficiently}
$$
	\begin{cases}
		\begin{aligned}
			\|\, \mathcal{F}\big( \hat{\bm{u}}(X),D^{\nabla}_{h}(X,Y)E^{\dagger}_{h}(Y,X)\hat{\bm{u}}(X),
			D^{\nabla^{2}}_{h}(X,Y)E^{\dagger}_{h}(Y,X)\hat{\bm{u}}(X),\dots;\btheta  \big) &- \bm{b}(X;\btheta) \,
			\|^{2}_{2}, \quad \x \in \Omega, \\
			\|\, \mathcal{G}\big( \hat{\bm{u}}(X),D^{\nabla}_{h}(X,Y)E^{\dagger}_{h}(Y,X)\hat{\bm{u}}(X),
			D^{\nabla^{2}}_{h}(X,Y)E^{\dagger}_{h}(Y,X)\hat{\bm{u}}(X),\dots;\btheta  \big) &- \bm{g}(X;\btheta) \,
			\|^{2}_{2},  \quad \x \in \partial\Omega.
		\end{aligned}
	\end{cases}
$$

By optimizing the above loss function for each physics-corrected parameter, we obtain the optimized parameter–correction pairs given by $\mathcal{D}_{\text{law}} = \{(\btheta_{\text{law},m}, \,\bm{\omega}_{m})\}_{m=1}^{M}$, with $\bm{\omega}_{m}=(\omega_{m1},\dots,\omega_{mK})^\top \in \mathbb{R}^K$. To further characterize the relationship between the entire parameter space and the correction coefficient space, we use $\mathcal{D}_{\text{law}}$ to learn $K$ independent RBF interpolation functions $s_{k}: \btheta \mapsto \omega_{k}$, similar to the GP surrogate, to approximate this mapping. The overall schematic of our LC-prior GP method is shown in Figure~\ref{fig:lc-prior_gp}.

\begin{algorithm}[t]
	\renewcommand{\algorithmicrequire}{\textbf{Input:}}
	\renewcommand{\algorithmicensure}{\textbf{Output:}}
	\caption{LC-prior GP}
	\label{likelihood}
	\begin{algorithmic}[1]
		\REQUIRE $\mathcal{D}_{\text{High}} = \{(\btheta_{n}, \bm{u}(\x;\btheta_{n}))\}_{n=1}^{N}$; the number of basis functions $K$; prediction target $\btheta^*$
		\ENSURE Surrogate approximate solution $\mathcal{M}_{\text{LC}}(\x;\btheta^*)$.
		\STATE Parameterize the output solutions in $\mathcal{D}_{\text{High}}$ by POD method and obtain the low-dimensional dataset $\mathcal{D}_{\text{low}} = \{(\btheta_{n}, \bm{\alpha}_{n})\}_{n=1}^{N}$.
		\STATE Construct GP surrogates for the $K$ basis coefficients $\alpha_k$: $f_{k}(\cdot) \sim \mathcal{GP}_{k}(\cdot)$ with dataset $\mathcal{D}_{\text{low}} $.
		\STATE Select $M$ physical correction parameters $\{\btheta_{\text{law}(m)} \}_{m=1}^{M} \in \Omega$.
		
		\FOR{$m = 1$ \textbf{to} $M$}
		\STATE \quad Predict $(\mu_{k}(\cdot), \sigma^{2}_{k}(\cdot))$ for each $\btheta_{\text{law}}$ using the GP surrogate $f_{k}(\cdot)$
		\STATE \quad In the bounds $[-z \sigma^{2}(\cdot), z \sigma^{2}(\cdot)]$, optimize the correction coefficient $\bm{\omega}_{k}$ using the physical law loss function in Eq. \eqref{eq:Loss_function}
		\ENDFOR
		\STATE Train a corrected model for each basis function coefficients $s_{k}(\cdot): \btheta \mapsto \bm{\omega}_{k}$ using the new training data $\mathcal{D}_{\text{law}} = \{(\btheta_{\text{law}(i)}, \bm{\omega}_{(m)})\}_{m=1}^{M}$ and interpolation functions.
		\STATE Renew the prior mean using $s_{k}(\cdot)$ to get the LC-prior GP: $\tilde{f}_{k}(\cdot) \sim \mathcal{N}(\tilde{\mu}_{k}(\cdot), \tilde{\sigma}^2_{k}(\cdot))$.
		\STATE Compute the approximate solution: $\mathcal{M}_{\text{LC}}(\x;\btheta^*) = \sum_{k=1}^{K} \tilde{\mu}_{k}(\btheta^*) \phi_{k}(\x)$, where the posterior mean $\tilde{\mu}_{k}(\btheta^*) = s_{k}(\btheta^*) + \mu_{k}(\btheta^*)$.
	\end{algorithmic}
	\label{Algorithm:lc-prior_gp}
\end{algorithm}

\subsection{Prediction of LC-prior GP}
\label{set:pre}
The complete surrogate consists of two parts: the data-driven GPR model $f_{k}: \btheta \mapsto \alpha_{k}$ and the physical law corrected model $s_{k}(\cdot): \btheta \mapsto \omega_{k} $, both models have the same inputs. Based on posterior formulation \eqref{eq:lc_posterior}, we can pass the prediction $s_{k}$ of the corrected model back to the GPR model as the physics constraints to renew the prior mean function in \eqref{eq:lc_prior}. So the LC-prior GP $\tilde{f}_{k}(\cdot)$ is also a Gaussian process with a law-corrected prior:
$$
    \tilde{f}_{k}(\btheta) \sim \mathcal{GP}\big( \tilde{m}_{k}(\btheta|\text{Law}) , \text{\bf{k}}_{k}(\btheta, \btheta^{'}) \big).
$$ 

Given a new parameter $\btheta^{*}$, the predicted posterior distribution can be written
$$
    \tilde{f}_{k}(\btheta^{*}) \sim \mathcal{N}(\tilde{\mu}_{k}(\btheta^*), \tilde{\sigma}^{2}_{k}(\btheta^*)),
$$
with the corresponding physical law-corrected posterior mean and std:
$$
\tilde{\mu}_{k}(\btheta^*) = \mathbb{E}[f_{k}(\btheta^*)|\btheta^*, \mathcal{D}_{\text{Low}}, \text{Law}] = s_k(\btheta^*) + \text{\bf{k}}_{k*}^\top \text{\bf{K}}^{-1}_{k} \bm{\alpha}_{k} = s_k(\btheta^*) + \mu_{k}(\btheta^*) ,
$$
$$
\tilde{\sigma}^{2}_{k}(\btheta^*) = \text{Var}[f_{k}(\btheta^*)|\btheta^*, \mathcal{D}_{\text{Low}},\text{Law}] = \text{\bf{k}}_{k}(\btheta^*, \btheta^*) - \text{\bf{k}}_{k*}^{\top} \text{\bf{K}}_{k}^{-1} \text{\bf{k}}_{k*}.
$$
And we can reconstruct the solution $\bm{u}(\x;\btheta^{*})$ in same domain by LC-prior GP
\begin{equation} \label{eq:M_lc}
	\begin{aligned}
		\mathcal{M}_{\text{LC}}(\x;\btheta^{*}) &= \sum^{K}_{k=1} \tilde{\mu}_{k}(\btheta^{*}) \phi_{k}(\x) 
							= \sum^{K}_{k=1} \big(\mu_{k}(\btheta^{*}) + s_{k}(\btheta^{*})  \big) \phi_{k}(\x) \\
							&= \sum^{K}_{k=1}  s_{k}(\btheta^{*}) \phi_{k}(\x) + \mathcal{M}_{\text{GP}}(\x;\btheta^{*}).
	\end{aligned}
\end{equation}

 {It is worth noting that the introduction of physical laws does not deteriorate predictive performance on the training data. Based on the derivation of the conditional posterior mean of the LC-prior GP, we observe that, compared with the standard GP, only an additional correction term is introduced, which is consistent with the prediction in Eq.~\eqref{eq:M_lc}. This demonstrates that the proposed framework is fully self-consistent in both formulation and logic. The overall framework of the LC-prior GP is summarized in Algorithm \ref{Algorithm:lc-prior_gp}. }

\subsection{Extension to multi-coupled PDE systems}
In many real-world scenarios, the governing equations consist of multi-coupled systems involving strongly interacting physical processes. %Extending the LC-prior GP framework to such settings requires a careful treatment of the correlations among multiple dependent variables. 
We consider the following general form of a multi-coupled PDE system:
\begin{equation}
	\label{eq:multi-pde}
	\begin{cases}
		\begin{aligned}
			 \mathcal{F} \big(\bm{u}^{(1)}, \bm{u}^{(2)}, \, \dots, \, \bm{u}^{(J)};\, \boldsymbol{\theta}\big)  = \bm{b}(\x;\btheta), \quad \boldsymbol{x} &\in \Omega, \\ 
			 \mathcal{G} \big(\bm{u}^{(1)}, \bm{u}^{(2)}, \, \dots, \, \bm{u}^{(J)};\, \boldsymbol{\theta}\big) = \bm{g}(\x;\btheta), \quad \boldsymbol{x}  &\in \partial\Omega.
		\end{aligned}
	\end{cases}
\end{equation}
The $\bm{u}^{(j)}(\x;\btheta)$ denotes the $j$-th physical field. Suppose training data $\{(\btheta_i, \bm{u}_n^{(1)}, \dots, \bm{u}_{n}^{(J)})\}_{n=1}^N$ have been obtained using the RBF-FD method. Following the same procedure as in the single-physics setting, each solution field is projected onto its corresponding reduced POD basis $\bm{\phi}^{(j)}$, yielding the coefficient vectors $\bm{\alpha}^{(j)}$ by  
$$
\text{POD}:\,\bm{u}^{(j)}(\x;\btheta_{n}) \in \mathbb{R}^D \mapsto \bm{\alpha}^{(j)}(\btheta_{n}) \in \mathbb{R}^{K^{(j)}}, \quad j=1,\dots,J,\quad n=1,\dots,N.
$$
The number of optimal POD modes $K^{(j)}$ is selected by cumulative energy \eqref{eq:pod_energy} for the $j$-th physical variable. For each coefficient, a GPR surrogate is trained independently
$$
 {f}^{(j)}_{k}(\btheta) \sim \mathcal{GP}\big( {m}^{(j)}_{k}(\btheta) , \text{\bf{k}}^{(j)}_{k}(\btheta, \btheta^{'}) \big), \quad j=1,\dots,J,\quad k=1,\dots,K^{(j)}.
$$
Combined with the fixed modes, a data-driven surrogate $\mathcal{M}_{\text{GP}}^{(j)}$ for each variable $\bm{u}^{(j)}(\x;\btheta)$ can be obtained.

A naive extension would treat these surrogates independently, but this ignores the cross-variable couplings encoded in the governing equations. To address this, the LC-prior GP framework is adapted by introducing joint correction coefficient $\omega_{k}^{(j)}(\btheta|\text{Law})$ for each GPR model, optimized simultaneously by Eq.~\eqref{eq:multi-pde}
\begin{equation}  
	\text{Loss} = || \mathcal{F}\big( \mathcal{M}_{\text{LC}}^{(1)},\dots,\mathcal{M}_{\text{LC}}^{(J)};\btheta \big) - \bm{b}(\x;\btheta) ||^{2}_{{2}} 
	+ \bm{\lambda} \cdot || \mathcal{G}\big( \mathcal{M}_{\text{LC}}^{(1)},\dots,\mathcal{M}_{\text{LC}}^{(J)};\btheta \big) - \bm{g}(\x;\btheta) ||^{2}_{{2}},
	\label{eq:multi_Loss_function}
\end{equation}
where $\mathcal{M}_{\text{LC}}^{(j)}$ denotes the surrogate reconstruction of the corresponding variable. Consistent with the previous formulation, we employ a small set of parameter samples to perform the optimization based on the loss function \eqref{eq:multi_Loss_function}, and learn the global correction mapping $s_{k}^{(j)}: \btheta \mapsto \omega^{(j)}_{k}$ for each surrogate model in the parameter space through interpolation functions. The interpolation functions are then back-propagated to the original GP surrogates as the LC-prior functions, yielding the LC-prior GP $\tilde{f}^{\,(j)}_{k}(\cdot)$. 

For given $\btheta^*$,  any physical variable $\bm{u}^{(j)}(\x;\btheta^*)$ can be efficiently predicted through the LC-prior GP
\begin{equation*}
\mathcal{M}^{(j)}_{\text{LC}}(\x;\btheta^*) = \sum^{K^{(j)}}_{k=1}  s^{(j)}_{k}(\btheta^{*}) \phi_{k}(\x) + \mathcal{M}^{(j)}_{\text{GP}}(\x;\btheta^{*}) ,\quad j=1,\cdots,J.
\end{equation*}

This joint optimization ensures that the correction terms are consistent across all variables and enforce the inter-dependencies dictated by the PDE system. In this way, the multi-coupled LC-prior GP explicitly encodes the physical couplings, thereby yielding predictions that are physically coherent and consistent.

\subsection{Parameter estimation by LC-prior GP}
\label{sec:p_estima}
Inferring unknown parameters \( \bm{\theta} \) from indirect observations \( \y \) is a critical application \cite{wang2021explicit}. Typically, the observed data \( \y \) is contaminated with noise, often modeled as \( \y=\mathcal{M}_{\text{true}}(\x;\btheta) + \bm{\epsilon} \), where \( \mathcal{M}_{\text{true}}(\x;\btheta) \) represents the true model output and \(\bm{\epsilon}\) is the noise term. There are two general frameworks for parameter estimation: (1) deterministic methods for point estimation, and (2) Bayesian inverse methods for posterior estimation \cite{andrieu2008tutorial}.
Here we propose to infer the unknown parameters in Bayesian framework with use of our LC-prior GP surrogate.

In Bayesian setting, the prior belief about the parameter $\bm{\theta}$ is encoded in the probability distribution $\pi_{\text{prior}}(\bbeta)$. Here we use a uniform distribution as prior for highlighting the action of likelihood function. Our aim is to infer the distribution of $\bbeta$ conditioned on the given data $\bm{y}$, the LC-prior GP surrogate \( \mathcal{M}_{\text{LC}} \) and physical constraints, the posterior distribution $\pi(\bbeta|\y, \mathcal{M}_{\text{LC}}, \text{Law})$. By the Bayes' rule, we have 
\begin{align}
	\pi(\bbeta|\y, \mathcal{M}_{\text{LC}}, \text{Law})&\propto P_{\epsilon} \big( \y - \mathcal{M}_{\text{LC}}(\x;\btheta) \big) \cdot \pi(\text{Law} | \mathcal{M}_{\text{LC}}, \y) \cdot \pi_{\text{prior}}(\bbeta),
	\label{eq:posterior}
\end{align}
where $P_{\epsilon} \big( \y - \mathcal{M}_{\text{LC}}(\x;\btheta) \big)$ is the likelihood $\pi(\y, \mathcal{M}_{\text{LC}}|\btheta)$ and $\pi(\text{Law} | \mathcal{M}_{\text{LC}}, \y)$ is the conditional distribution of physical law defined by the loss function Eq. \eqref{eq:Loss_function}
$$P_{\epsilon} \big( \y - \mathcal{M}_{\text{LC}}(\x;\btheta) \big) \propto \exp(-||y-\mathcal{M}_{\text{LC}}(\x;\theta)||^{2}_{2} ),$$
$$\pi(\text{Law}|\mathcal{M}_{\text{LC}},\y) \propto \exp(-||\text{Loss} ||_{2}^{2}),$$
fidelity to the DEs can be measured by $\pi(\text{Law}|\mathcal{M}_{\text{LC}},\y)$, while the data can be measured by 
$P_{\epsilon} \big( \y - \mathcal{M}_{\text{LC}}(\x;\btheta) \big)$. Weights for observations and equation residual are assumed to be the same.

The unnormalized posterior Eq.~\eqref{eq:posterior} can be efficiently sampled using Metropolis–Hastings (MH) algorithm \cite{chib1995understanding}, a useful Markov Chain Monte Carlo (MCMC) method \cite{andrieu2008tutorial}.
An MH step with invariant distribution $\pi(\btheta)$ and proposal distribution $\pi_q(\btheta' \mid \btheta^{(i)})$ proceeds by generating a candidate $\btheta'$ from $\pi_q(\btheta' \mid \btheta^{(i)})$ given the current state $\btheta^{(i)}$. The candidate is accepted with probability
\begin{equation} \label{eq:acceptance_p}
\mathcal{A}(\bbeta^{(i)},\bm{\theta}^{'})=\min \left\{1,\frac{\pi(\bbeta^{'}|\y)q(\bbeta^{(i)}|\bbeta^{'})}{\pi(\bbeta^{(i)}|\y)q(\bbeta^{'}|\bbeta^{(i)})} \right\} ,
\end{equation}
otherwise it remains at $\btheta^{(i)}$
In our work, we draw from $\pi(\bbeta|\y, \mathcal{M}_{\text{LC}})$ in Eq.~\ref{eq:posterior} using MH sampling:
\begin{enumerate}
	\item initialize $\bm{\theta}^{1}$.
	\item For $i$ = 1 to $N$: \\ 
	(a) Sample $\btheta^{'}$ from the proposal distribution $\pi_q(\bm{\theta}^{'}|\bm{\theta}^{(i)})$.\\
	(b) Compute $\mathcal{A}(\bbeta^{(i)},\bm{\theta}^{'})$ by Eq.~\eqref{eq:acceptance_p} and sample a constant $a$ from $\mathcal{U}[0,1]$.\\
	(c) If $a < \mathcal{A}(\bbeta^{(i)},\bm{\theta}^{'})$ then accept $\bbeta^{'}$.  Otherwise, set $\bbeta^{'}=\bbeta^{(i)}$.
\end{enumerate}

Bayesian methods offer advantages by explicitly accounting for parameter uncertainty, making them more robust in the presence of noisy or sparse data. Although the sampling process in forward solving can be computationally intensive, this issue can be effectively alleviated by our proposed surrogate \( \mathcal{M}_{\text{LC}}\).

\section{Numerical examples}
\label{sec:numerical examples}
 {To validate the performance of the LC-prior GP method, we present numerical results evaluating the surrogate accuracy of the proposed approach and compare it with the standard GP method and the DMD-wiNN method \cite{song2024model}. Additionally, within the framework of our surrogate model, we demonstrate parameter estimation applications in two examples. 
And all experiments are based on a small-sample setting, where only 2–3 training points per parameter dimension are used to generate the training data.
To assess model accuracy, all subsequent experiments report the relative $L^1$ error defined in Eq.~\eqref{eq:norm1_error}. The spatial discretization of the RBF-FD scheme is constructed on point clouds generated by DistMesh \cite{persson2004simple}, with the nodal spacing $h$ specified in each subsection.}

\subsection{Reaction-diffusion model}
\begin{figure}[t!]
	\centering
	\includegraphics[width=1\linewidth]{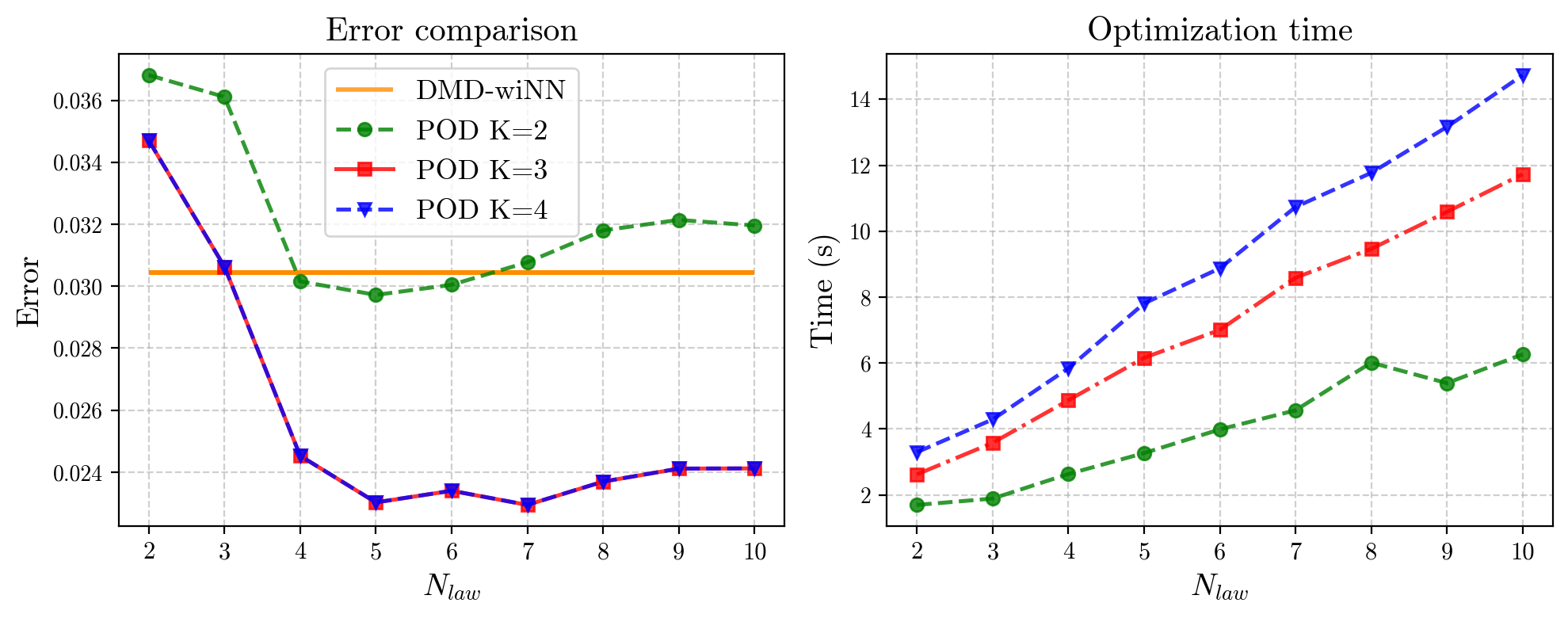}
	\caption{ {Relative errors obtained by the LC-prior GP with different $N_{\text{law}}$ and POD modes $K$ (left), and the optimization time with physical law correction under the corresponding conditions (right).}}
	\label{fig:AC_POD_k}
\end{figure}

In this subsection, we consider a reaction–diffusion equation \cite{kondo2010reaction} with homogeneous Dirichlet boundary conditions to provide a few basic verifications of the LC-prior GP. The governing equation reads
\begin{equation}
	\begin{cases}
		\bm{u}_t = \epsilon^2 \Delta u - F'(\bm{u}) + f(x, y, t), &\quad \text{in } [0, T] \times \Omega, \\
		F(u) = \frac{1}{4} (\bm{u}^2 - 1)^2 , &\quad \text{in } [0, T] \times \Omega, \\
		\bm{u}(\cdot,0) = \bm{u}_0,  &\quad \text{in }  \Omega.
	\end{cases}
	\label{eq:Reaction-diffusion model} 
\end{equation}
here we set $T=1$, $\Omega=[-1,1]^2$ and $f(x, y, t)=0$. The Eq.~\eqref{eq:Reaction-diffusion model} now is the classical Allen-Cahn equation, which involves a parameter $\epsilon$ related to the interface thickness. The initial value $\bm{u}_0$ is given by
\[
\bm{u}_0 = 
\begin{cases} 
	1, & \text{if } (x^2 + y^2)^{\frac{1}{2}} \leq \frac{1}{8}\left(3 + 3\sin(5\gamma)\right), \\
	0, & \text{elsewhere},
\end{cases} \quad \text{with} \,
\gamma = 
\begin{cases}
	\arccos\left(\frac{x}{\sqrt{x^2 + y^2}}\right), & y \geq 0, \\
	2\pi - \arccos\left(\frac{x}{\sqrt{x^2 + y^2}}\right), & y < 0.
\end{cases}
\]
 {The discrete scheme is given by $\bm{u}^{n+1} - \tau \epsilon^2 \Delta \bm{u}^{n+1} = \bm{u}^n - \tau f(\bm{u}^n)$ with $\tau=0.1$, and nodal spacing $h = 0.025$.}

Here $\btheta = \epsilon$. The training data consists of three sample parameters, $\epsilon = \{0, 0.05, 0.1\}$, used for forward simulations, and 200 samples are randomly drawn from the uniform distribution $\pi_{\text{prior}}(\epsilon) \sim \mathcal{U}[0,0.1]$ to generate the test data.
In this numerical example, only the surrogate model for ${u}(x,y,t;\btheta)$ needs to be constructed.

In this section, we choose the number of POD modes $K \in \{2,3,4\}$ and the number of physical law-corrected points $N_{\text{law}} \in \{2,\dots,10\}$ to investigate their impact on the LC-prior GP. We compare the relative prediction errors under different settings, as shown in Figure~\ref{fig:AC_POD_k} (left).  {Based on Eq.~\ref{eq:pod_energy}, we compute the cumulative energy ratio. When $K=2$, $\eta(2)=97.68\%$, which is insufficient to provide an accurate approximation of the solution space, leading to model failure. In contrast, when $K=3$ and $K=4$, $\eta(K) > 99.99\%$, resulting in satisfactory correction performance. It is worth noting that once the approximation capability is sufficient, further increasing the number of modes does not improve model performance. Therefore, following this criterion, we select the smallest $K$ that satisfies $\eta(K) > 99.99\%$ to balance accuracy and efficiency. We observe that when $N_{\text{law}}$ is approximately twice the size of the training data, the prediction error gradually stabilizes and reaches satisfactory accuracy with relatively low computational cost. Accordingly, we set $N_{\text{law}} = 7$ and $K = 3$ for constructing the surrogate model.}

\begin{table}[t]
	\centering
	\begin{tabular}{lccc}
		\toprule
		&  &  LC-prior GP & PI-DeepONet (parametric setting) \\
		\midrule
		&  Error     & $\bf{0.0230}$  &  $0.8465$ \\
        &  Training time (s)    & $8.59$  &  $830.73$ \\
		\bottomrule
	\end{tabular}
	\caption{ {The relative errors and training cost of the LC-prior GP and PI-DeepONet (parametric setting).}}
	\label{tab:AC_error}
\end{table}

\begin{figure}[t]
	\centering
	\includegraphics[width=1\linewidth]{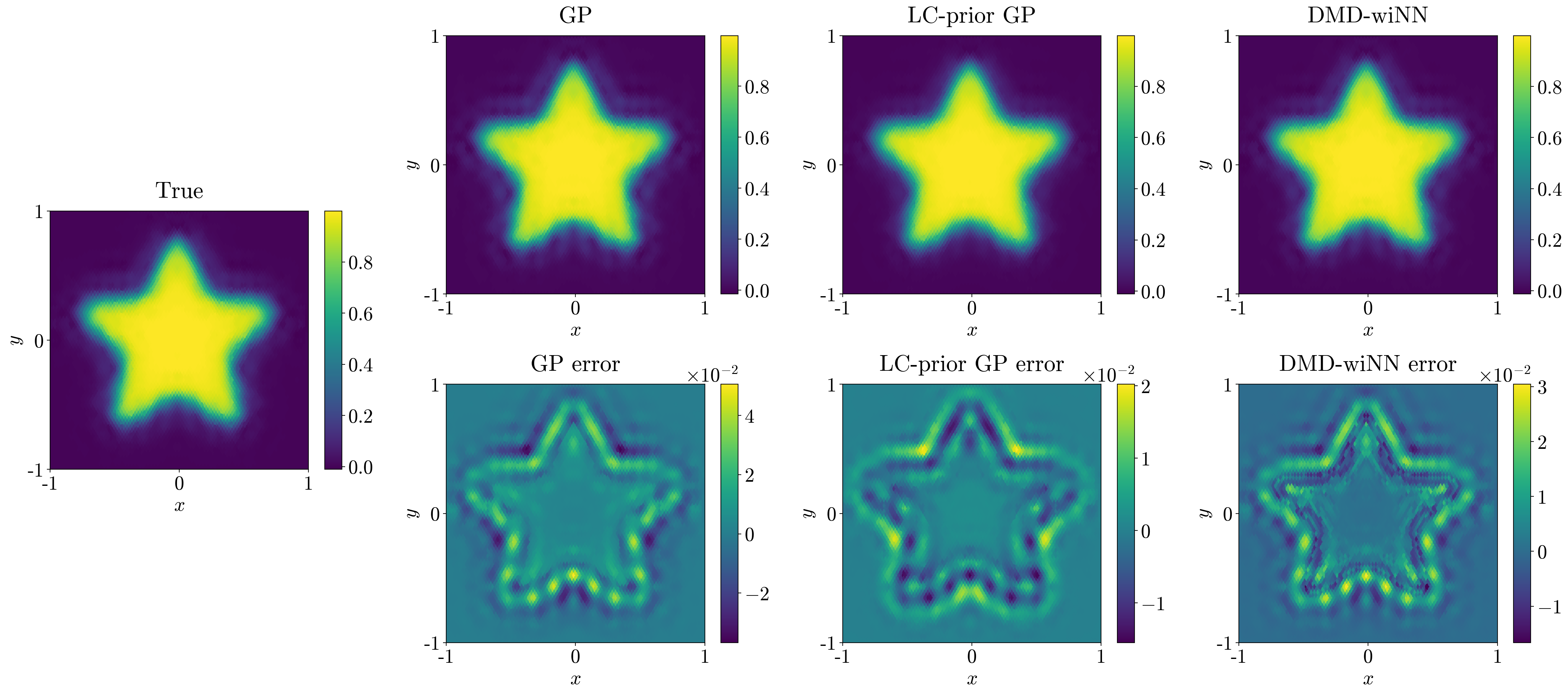}
	\caption{Results for the reaction–diffusion model: mean predictions of $u(x,y,t;\btheta)$ at $t=1$ by different methods and the corresponding pointwise errors.}
	\label{fig:AC_T1}
\end{figure}

 {To highlight the advantages in reduced sample requirements and the efficiency of physical correction, we conduct a fair comparison with DNN-based physics-informed methods. We adopt the network architecture of PI-DeepONet \cite{wang2021learning}, using a $70 \times 3$ architecture for both the branch and trunk networks. Note that in our setting, the branch input degenerates to a scalar parameter rather than a discretized input function, resulting in a parametric mapping instead of a standard operator learning problem, and the number of epochs is set to 1000 to ensure convergence. The same training data are adopted as in LC-prior GP, i.e., 3 samples for the data-driven mean square loss and 7 samples for the physics-informed loss. Table~\ref{tab:AC_error} reports the relative $L^1$ errors over the test data as well as the detailed training time. The PI-DeepONet fails to provide accurate predictions under such limited samples, while the LC-prior GP, based on physics-correction and differentiation matrices, exhibits more competitive results.}

 {Considering that DNN-based methods generally struggle under small-sample regimes, we only include the DMD-wiNN method \cite{song2024model}, which is a novel reduced-basis method designed for limited data scenarios, as the baseline. Figure~\ref{fig:AC_T1} presents the mean predictions at $t=1$ along with the corresponding pointwise errors. The overall error is $0.0839$ for the standard GP and $0.0307$ for DMD-wiNN. The results demonstrate the effectiveness of the LC-prior GP in learning physical constraints and achieving strong predictive performance.}

\subsection{Advection equation}
 {In this section, we extend our framework to a non-diffusion advection equation \cite{ewing2001summary} with Dirichlet boundary condition posed on an irregular domain:
\begin{equation}
	\begin{cases}
		\bm{u}_t - \beta \cdot \nabla \bm{u} = 0, &\quad \text{in } [0, T] \times \Omega, \\
		\bm{u}(\cdot,0) = \bm{u}_0,  &\quad \text{in } \Omega.
	\end{cases}
	\label{eq:advection}
\end{equation}
Here, $T = 1$ and $\Omega = [-1,1]^2 \setminus B_r(0)$, with $B_r(0) = \{ (x,y) \in \mathbb{R}^2 : x^2 + y^2 < r^2 \}$ and $r=0.4$ representing a classical perforated domain. The initial condition is defined as: ${u}_0 = \cos\left(\frac{\pi x}{2}\right)\sin\left(\frac{\pi y}{2}\right)$. The discrete scheme we used is $\bm{u}^{n+1} + \tau  c \nabla \bm{\phi}^{n+1} = \bm{u}^n$ with $\tau=0.1$, and nodal spacing $h = 0.03$.

\begin{table}[t]
	\centering
	\begin{tabular}{lcccc}
		\toprule
		&  &  $m_k^{(\mathrm{0})}(\btheta)\equiv 0$ & $m_k^{(\mathrm{C})}(\btheta) \equiv \mathbb{E}(\bm{\alpha}_k)$ & $m_k^{(\mathrm{L})}(\btheta)= \bm{a}^\top\btheta + b$  \\
		\midrule
		&   GP         & 0.1215  & 0.1211  & 0.4695 \\
        & LC-prior GP  & \textbf{0.0563}  & 0.0590  & 0.0802 \\
		\bottomrule
	\end{tabular}
	\caption{ {The relative errors of GP method and LC-prior GP with different prior mean functions.}}
	\label{tab:advection}
\end{table}

\begin{figure}[t]
	\centering
	\includegraphics[width=1\linewidth]{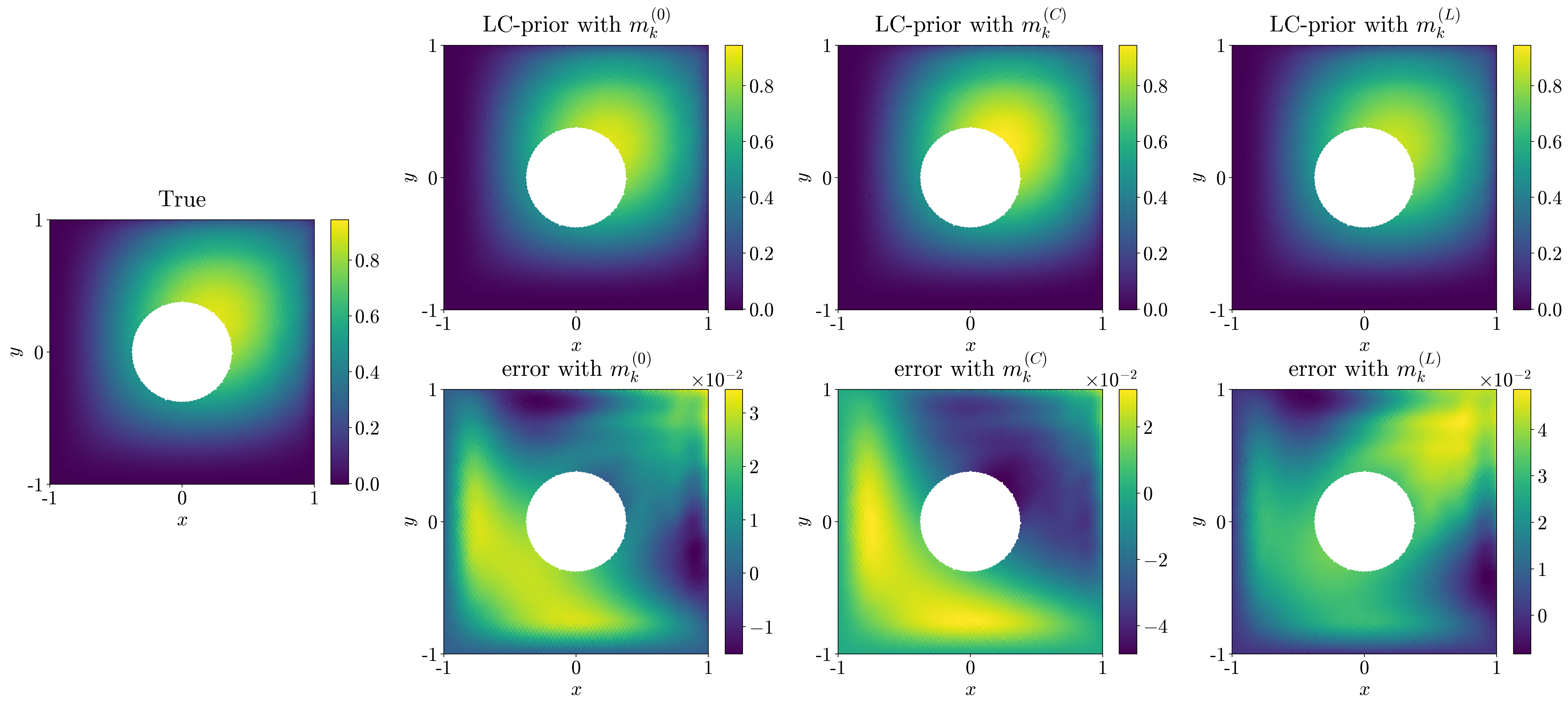}
	\caption{ {Results for the advection model: mean predictions of $u(x,y,t;\btheta)$ at $t=1$ by LC-prior GP with different prior mean function and the corresponding pointwise errors.}}
	\label{fig:advection}
\end{figure}

Here, $\btheta = \beta$. Only two parameter values $\beta = \{0.167,\,0.32\}$, are used to generate the training data. Furthermore, 200 test data are randomly drawn from the $\pi_{\text{prior}}(\beta) \sim \mathcal{U}[0,0.5]$ to evaluate the extrapolation performance of the LC-prior GP. Due to the extremely limited training data and the presence of out-of-distribution test cases, 10 physics-corrected parameters are selected from $\pi_{\text{prior}}(\beta)$ in this scenario to ensure sufficient coverage for the correction. The RB model is constructed using $K = 4$ modes.

In this section, we investigate the impact of the prior mean function $m_k(\btheta)$ on the subsequent modeling performance. We consider three representative choices. The first is the standard zero-mean assumption 
$$m_k^{(\mathrm{0})}(\btheta) \equiv 0.$$ 
The second adopts a constant, data-driven mean given by the empirical average of the coefficient samples, 
$$m_k^{(\mathrm{C})}(\btheta) \equiv \mathbb{E}(\bm{\alpha}_k).$$ 
The third introduces a parameterized linear trend of the form 
$$m_k^{(\mathrm{L})}(\btheta)= \bm{a}^\top\btheta + b,$$ 
where the $\bm{a}=(a_1,\dots,a_q)^\top\in\mathbb{R}^q$ and $b$ are jointly optimized together with the kernel hyperparameters. 

Based on the above three assumptions, we construct the corresponding LC-prior GP with kernel hyperparameters optimized separately. Table~\ref{tab:advection} reports the detailed errors over the test dataset under each setting. It is observed that both $m_k^{(0)}(\btheta)$ and $m_k^{(\mathrm{C})}(\btheta)$ correspond to constant prior means, so the predictions on the test set are primarily governed by the kernel function, leading to comparable accuracy in these two cases. In contrast, under the linear prior $m_k^{(\mathrm{L})}(\btheta)$, the GP predictions in extrapolation regions are dominated by the prior mean function, since the imposed linear trend is not consistent with the true underlying mapping. The second row reports the physics-corrected results, where all settings exhibit improved performance and outperform the $0.0814$ error by DMD-wiNN. However, $m_k^{(\mathrm{L})}(\btheta)$ remains limited by the bias in the prior mean. Even after correction within the 95\% confidence interval, its accuracy is still inferior to the other two settings.

Figure~\ref{fig:advection} visualizes the predicted mean solutions of the LC-prior GP. The results indicate that, although the physics-based correction is effective under all three prior mean settings, the choice $m_k^{(\mathrm{L})}(\btheta)$ tends to make the GP overly reliant on a potentially misspecified prior mean during prediction. Therefore, in the subsequent experiments, we adopt the assumption $m_k^{(0)}(\btheta) \equiv 0$, which avoids introducing prior bias on the test data and allows the model to focus on learning through the kernel function and the physical constraints. A detailed analysis is provided in Appendix \ref{sec:non_zero_prior}.

}

\subsection{Incompressible miscible flooding model}
In this subsection, we test a model with multiple parameters. The incompressible miscible flooding in the porous media is widely used in the engineering fields such as the reservoir simulation and the exploration of the underground water and oil. The classical formulations are given as \cite{song2024model}
\begin{equation}
	\begin{cases}
		\nabla \cdot \bm{u} = q, & \text{in } [0, T] \times \Omega, \\
		\bm{u} = -\frac{\kappa}{\mu(c)} \nabla p, & \text{in } [0, T] \times \Omega, \\
		\phi c_t + \bm{u} \cdot \nabla c = \nabla \cdot (\mathbf{D}(\bm{u}) \nabla c), & \text{in } [0, T] \times \Omega,
	\end{cases}
	\label{eq:flooding_1}
\end{equation}
where \(\Omega \in \mathbb{R}^2\) . The parameter \(\kappa\) represents the permeability, \(\mu\) represents the viscosity, and \(\phi\) is the porosity. The unknown functions \(\bm{u}\), \(p\) and \(c\) are the velocity, pressure, and concentration, respectively.

By replacing the velocity \(\bm{u}\) with the pressure \(p\), Eq.\eqref{eq:flooding_1} is equivalent to:
\begin{equation}
	\begin{cases}
		-\frac{\kappa}{\mu(c)} \Delta p = q, & \text{in } [0, T] \times \Omega, \\
		\phi c_t - \frac{\kappa}{\mu(c)} \nabla p \cdot \nabla c = \nabla \cdot (\mathbf{D}(\bm{u}) \nabla c), & \text{in } [0, T] \times \Omega, \\
        \mathbf{D}(\bm{u}) = d_m I + |\bm{u}| \big(d_l E(\bm{u}) + d_t (I - E(\bm{u}))\big), & \text{in } [0, T] \times \Omega,
	\end{cases}
	\label{eq:flooding_2}
\end{equation}
where \(I\) is the identity matrix, \(d_m\) is the effective diffusion coefficient, \(d_l\) is the longitudinal dispersion coefficient, \(d_t\) is the transverse dispersion coefficient, and \(E(\bm{u})\) is the projection tensor follows:
\[
(E(\bm{u}))_{i,j} = \frac{u_i u_j}{|\bm{u}|^2}, \quad 1 \leq i, j \leq d.
\]

\begin{figure}[t]
	\centering
	\includegraphics[width=1\linewidth]{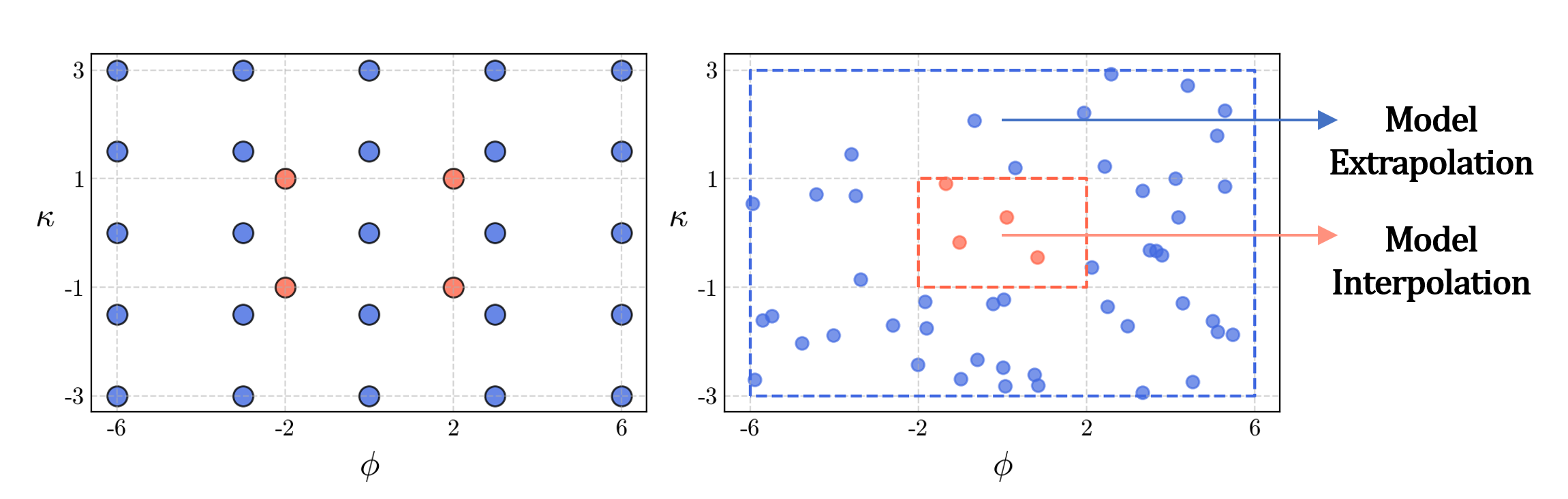}
	\caption{ {The strategy for selecting training and physics correction parameters (left) and some test data for a brief illustration (right). Blue points: $\btheta_{\text{obs}}$ used to train $f_k(\cdot)$; Orange points: $\btheta_{\text{law}}$ for physical law correction.}}
	\label{fig:flooding_p2_point}
\end{figure}

\begin{figure}[t!]
	\centering
	\includegraphics[width=1\linewidth]{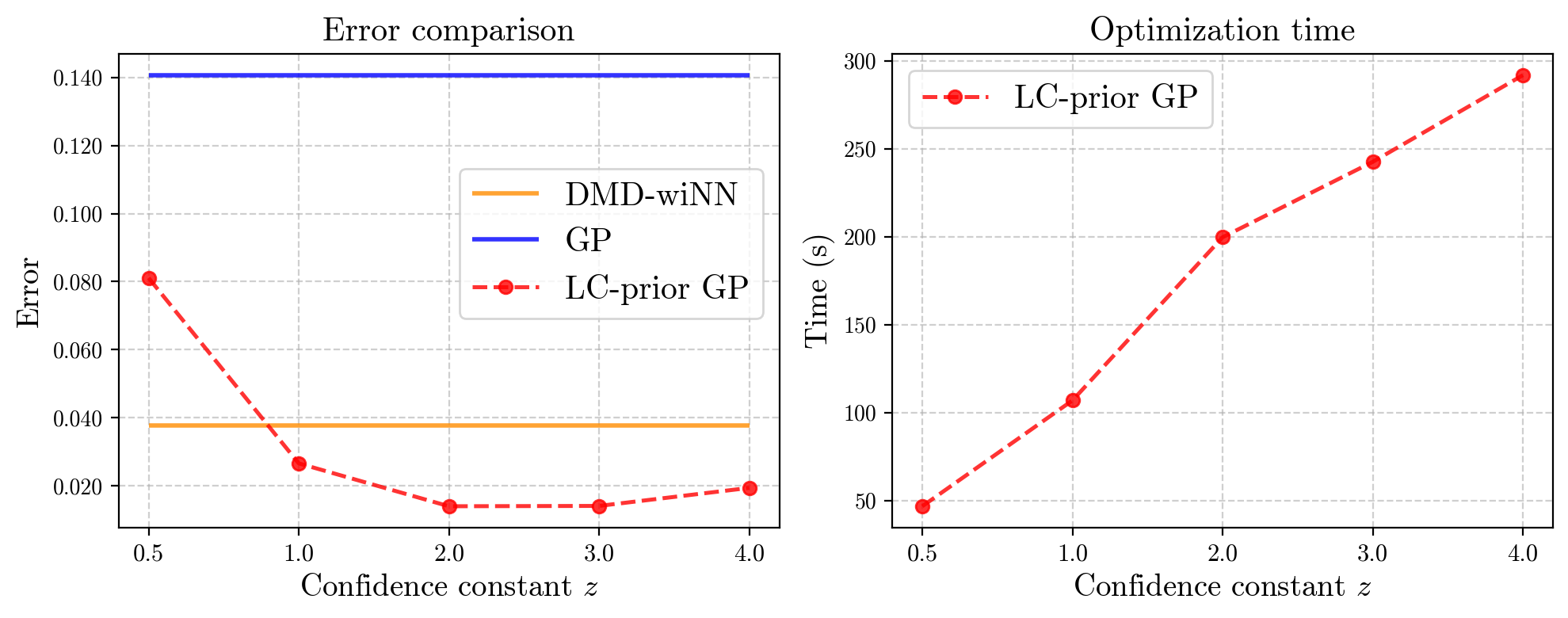}
	\caption{ {Results for the miscible flooding model: relative errors of LC-prior GP optimized by different confidence constant $z$ (left), and the optimization time under the corresponding conditions (right).}}
	\label{fig:flooding_p2_c}
\end{figure}

For this example, we consider \(\mathbf{D}(\bm{u}) = d_m I\) and set \(T = 0.1\). We select an irregular region in the two-dimensional space \(\Omega\). The radius of this region satisfies the following requirement:
\[
	r_a = 1 + \frac{\sin(7\gamma) + \sin(\gamma)}{10}, \quad \gamma \in [0, 2\pi].
\]
The $\Omega$ is bounded by \([-1.5, 1.5]^2 \), and the distance after the division is \(h = 0.04\).  {The discrete numerical scheme of the temporal direction with $\tau=0.01$ is as follows:}
\[
	\begin{cases}
		-\frac{\kappa}{\mu(c)} \Delta p^{n+1} = q^{n+1}, \\
		\phi \frac{c^{n+1} - c^n}{\tau} - \frac{\kappa}{\mu(c)} \nabla p^n \cdot \nabla c^{n+1} - d_m \Delta c^{n+1} = 0.
	\end{cases}
\]

\subsubsection{Two parameters example}
\label{sec:two-params}
Here $\btheta = (\kappa, \phi)$. The training data is selected by $\kappa = \{-1, 1\}$ from $\pi_{\text{prior}}(\kappa) \sim \mathcal{U}[-3,3]$ and $\phi = \{-2, 2\}$ from $\pi_{\text{prior}}(\phi) \sim \mathcal{U}[-6,6]$, yielding $4$ training data from their Cartesian product to examine whether applying physics law corrections beyond the training data can further improve the LC-prior GP's extrapolation performance on the test set, while the test data contains 400 samples randomly scattered in the same distribution. For the physics corrected set, 5 samples equally spaced points were sampled from each prior distribution, yielding a total of 25 test data. Figure~\ref{fig:flooding_p2_point} shows a concise schematic representation. We construct surrogate models for both $p(x,y,t;\btheta)$ and $c(x,y,t;\btheta)$ simultaneously to enable subsequent correction through the loss function \eqref{eq:multi_Loss_function}.  {During the parametric representation, the $K=3$ modes are selected for both two variables.} 

\begin{table}[t]
	\centering
	\begin{tabular}{lcccc}
		\toprule
		&  & GP & LC-prior GP & DMD-wiNN  \\
		\midrule
		& Two parameters example    & $0.1407$  & $\bf{0.0140}$  & $0.0377$ \\
		& Three parameters example  & $0.0528$  & $\bf{0.0100}$  & $0.0103$ \\
		\bottomrule
	\end{tabular}
	\caption{ {The relative errors of the GP method, the LC-prior GP method, and DMD-wiNN.}}
	\label{tab:flooding_p2}
\end{table}

\begin{figure}[t]
	\centering
	\includegraphics[width=1\linewidth]{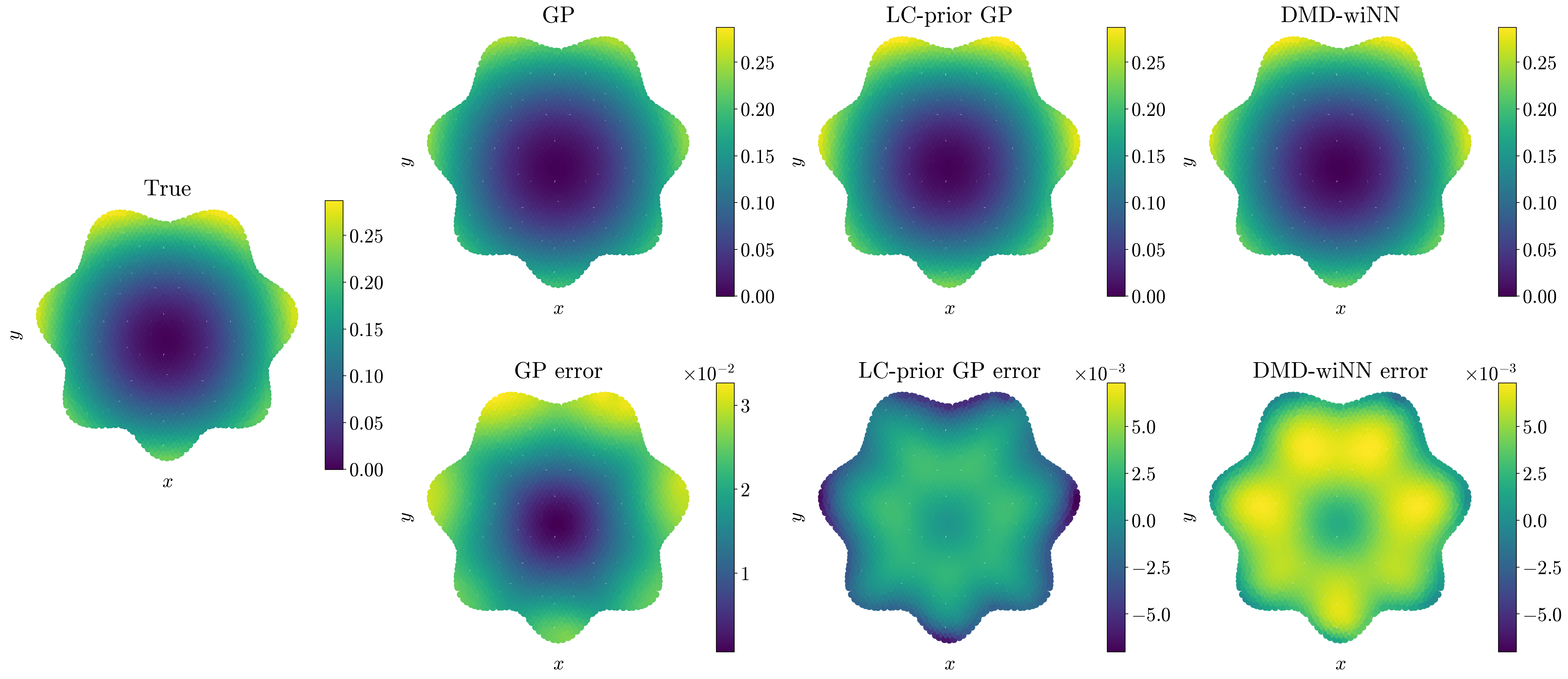}
	\caption{Results for the two parameters miscible flooding model: mean predictions of $c(x,y,t;\btheta)$ at $t=0.1$ by different methods and the corresponding pointwise errors.}
	\label{fig:flooding_p2}
\end{figure}

\begin{table}[t]
	\centering
	\begin{tabular}{lcc}
		\toprule
		& Method & (Mean ± Std) of posterior \\
		\midrule
		$\sigma_{\text{obs}}^2=0.1$ & LC-prior GP  & \textbf{(0.68 ± 0.31, 5.12 ± 0.63)}  \\
		& GP           & (2.21 ± 0.54, 5.30 ± 0.55)           \\
		\hline
		$\sigma_{\text{obs}}^2=0.2$ & LC-prior GP  & \textbf{(0.98 ± 0.33, 4.72 ± 0.48)}  \\
		& GP           & (1.55 ± 0.51, 3.90 ± 0.35)           \\
		\bottomrule
	\end{tabular}
	\caption{Posterior samples obtained by MH for the two parameter example with $\btheta^{*}=(0.64,,4.97)$.}
	\label{tab:flooding_p2_mcmc}
\end{table}

 {First, we consider confidence constants $z \in \{0.5,1,2,3,4\}$ in the interval $[\mu-z\sigma,\,\mu+z\sigma]$ to investigate the effect of the confidence interval on enforcing physical constraints. Figure~\ref{fig:flooding_p2_c} shows the relative $L^1$ error of $\hat{c}(x,y,t;\btheta)$ over 400 test data points for different values of $z$. It is observed that the model achieves the best accuracy when $z=2$. Further enlarging the optimization interval does not improve performance; instead, it increases computational cost. Therefore, selecting the 95\% confidence interval strikes a good balance between accuracy and efficiency, and we set $z=2$ for all subsequent experiments.}  
Figure~\ref{fig:flooding_p2} visualizes the mean predictions of $\hat{c}(x,y,t;\btheta)$ at $t=0.1$, comparing the performance of the three methods. Table~\ref{tab:flooding_p2} reports the detailed relative $L^1$ errors. It is noteworthy that the LC-prior GP achieves the best performance, improving the accuracy by approximately one order of magnitude compared to the standard GP, with errors of 0.0154 and 0.1407, respectively. This clearly demonstrates the effectiveness of the physical law correction.

In the parameter estimation applications, to further evaluate the model's robustness, we randomly selected a test sample with $\btheta^{*}=(0.64,4.97)$ and added white Gaussian noise at varying intensity levels to simulate observed measurements. The MH sampling in this section follows the process introduced in Section~\ref{sec:p_estima}. During sampling, the number of iterations is set to 10,000, with the first 1,000 samples discarded as burn-in. The posterior mean and standard deviation are presented in Table~\ref{tab:flooding_p2_mcmc}. The posterior samples based on the LC-prior GP surrogate demonstrate more accurate mean estimates compared to those from the standard GP surrogate under both noise conditions.

\subsubsection{Three parameters example}

\begin{figure}[t!]
	\centering
	\includegraphics[width=1\linewidth]{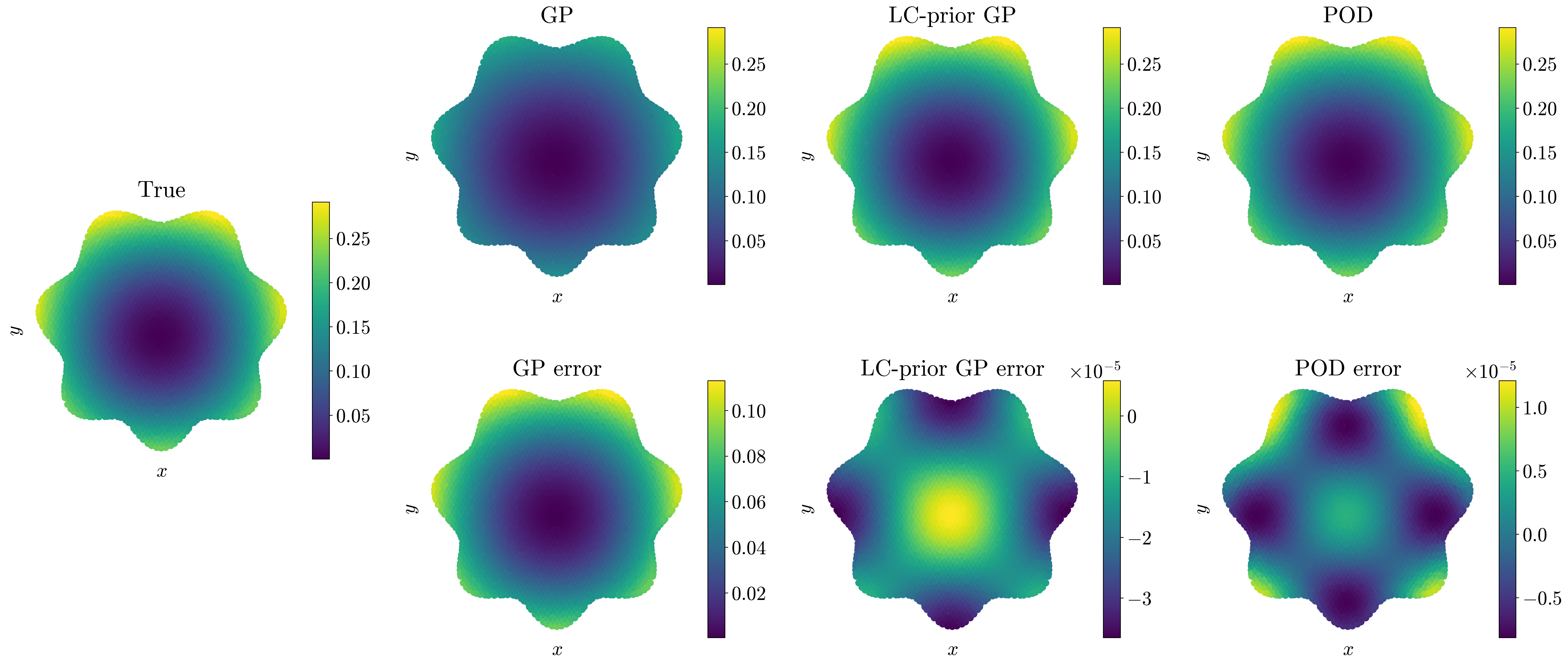}
	\caption{ {Results for the three parameter miscible flooding model: predictions for a test case using GP and LC-prior GP, the exact reduced representation by POD, and the corresponding pointwise errors.}}
	\label{fig:flooding_p3_optima}
\end{figure}

In this subsection, we further test the parametric PDE based on Eq.\eqref{eq:flooding_2} and extend to the three parameters example as $\btheta = (\kappa, \mu, \phi)$. The size of the training data is $8$ with evenly scattered $2$ points in intervals $\pi_{\text{prior}}(\kappa) \sim \mathcal{U}[-3,3]$, $\pi_{\text{prior}}(\phi) \sim \mathcal{U}[-6,6]$ and $\pi_{\text{prior}}(\mu) \sim \mathcal{U}[-10,10]$. The size of the test data is $8000$ with randomly sampled and size of the physics corrected points is $64$ with evenly scattered $4$ interior points, excluding the boundary points in prior distribution. The discretization scheme remains the same as in two parameters example.  {The leading 3 POD modes is chosen independently by Eq.~\eqref{eq:pod_energy} for both $p(x,y,t;\btheta)$ and $c(x,y,t;\btheta)$.}

\begin{table}[t!]
	\centering
	\begin{tabular}{lcc}
		\toprule
		& Method & (Mean ± Std) of posterior \\
		\midrule
		$\sigma_{\text{obs}}^2=0.1$ & LC-prior GP  & \textbf{(1.59 ± 0.46, -2.41 ±  0.70, -4.47 ± 1.15)}  \\
		& GP           & (1.31 ± 0.25, -1.08 ± 0.65, -1.57 ± 1.30)           \\
		\hline
		$\sigma_{\text{obs}}^2=0.2$ & LC-prior GP  & \textbf{(1.78 ± 0.26, -2.01 ± 0.27, -4.72 ± 0.43)}  \\
		& GP           & (0.91 ± 0.28, 2.83 ± 0.49, -5.58 ± 0.62)           \\
		\bottomrule
	\end{tabular}
	\caption{Posterior samples obtained by MH for the three parameters example with $\btheta^{*}=(1.45,-2.21,-4.68)$}
	\label{tab:flooding_p3_mcmc}
\end{table}

\begin{figure}[t]
	\centering
	\includegraphics[width=1\linewidth]{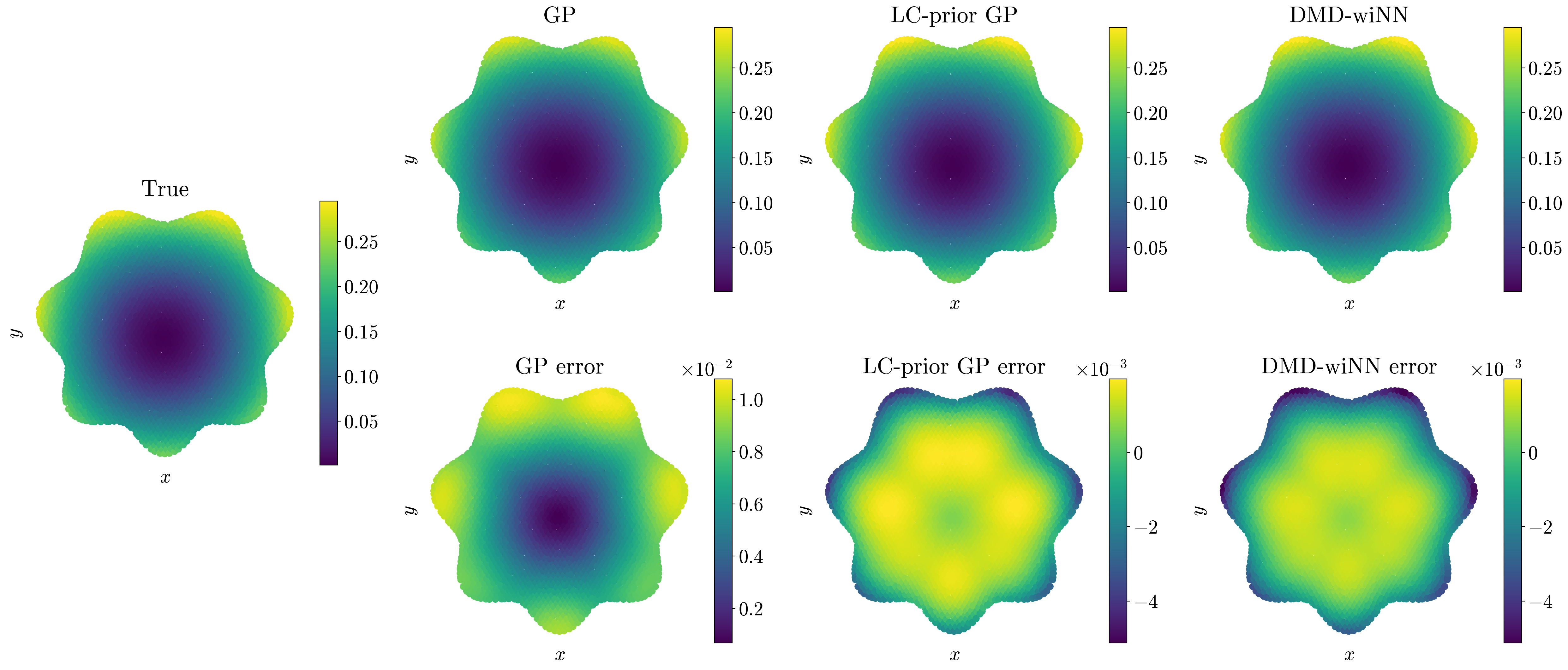}
	\caption{Results for the three parameters miscible flooding model: mean predictions of $c(x,y,t;\btheta)$ at $t=0.1$ by different methods and the corresponding pointwise errors.}
	\label{fig:flooding_p3}
\end{figure}

To demonstrate the effectiveness of the physical correction, we selected the sample with the largest error in the physics corrected points $\btheta_{\text{law}}$ and visualized its correction performance. Figure~\ref{fig:flooding_p3_optima} presents the detailed results of $\hat{c}(x,y,t)$. It can be clearly observed that, after correction, the pointwise error $\hat{c}-c$ is reduced by approximately four orders of magnitude (from $10^{-1}$ to $10^{-5}$), approaching the theoretical accuracy limit of the reduced-order model (fourth column). The mean solutions of the concentration $c$ and the corresponding relative errors on the test set are shown in Figure~\ref{fig:flooding_p3} and Table~\ref{tab:flooding_p2}, respectively. Moreover, the proposed method consistently achieves the best performance for multi-parameter systems defined on irregular domains with smooth boundaries.

The parameter estimation in this example, we follow the same workflow as described in the two parameters example with the number of iterations is set to 10000 and the first 1000 samples discarded as burn-in. The detailed posterior results presented in the accompanying Table~\ref{tab:flooding_p3_mcmc}. Under our proposed framework, the posterior samples for systems with multi-physics coupling and multiple parameters demonstrate more competitive performance compared to purely data-driven surrogate models.

\subsection{Incompressible Navier-Stokes model}
In this subsection, we mainly consider the incompressible Navier-Stokes model which can be used to describe many fluid phenomenons \cite{boyer2012mathematical}. 
The incompressible Navier-Stokes model with the Dirichlet boundary condition is as follows:
\begin{table}[t]
	\centering
	\begin{tabular}{lcccc}
		\toprule
		&  & GP & LC-prior GP & DMD-wiNN  \\
		\midrule
		& $\bm{u}$ in the x-direction  & $0.0309$  & $\bf{0.0202}$  & $0.0320$ \\
		& $\bm{u}$ in the y-direction  & $0.0265$  & $\bf{0.0163}$  & $0.0264$ \\
		\bottomrule
	\end{tabular}
	\caption{The relative errors of the GP method, the LC-prior GP method, and DMD-wiNN.}
	\label{tab:ns_error}
\end{table}

\begin{equation}
	\begin{cases}
		\bm{u}_t + (\bm{u} \cdot \nabla)\bm{u} - \mu\Delta\bm{u} + \frac{\nabla p}{\rho} = 0, & \text{in } [0, T] \times \Omega, \\
		\nabla \cdot \bm{u} = 0, & \text{in } [0, T] \times \Omega, \\
		\bm{u}(\cdot,0) = \bm{u}_0, & \text{in } \Omega,
	\end{cases}
	\label{eq:ns}
\end{equation}
where $\Omega\in \mathbb{R}^2$ is selected an irregular region that satisfies the same requirement in Eq.\eqref{eq:flooding_1} and $T = 0.1$. The $\bm{u}=[u,v]'$ represents the velocity components in the x- and y-directions, $\mu$ is the viscosity, $p$ is the pressure, and $\rho$ is the density. The initial value is given by: 
$ \bm{u}_0(x, y) = \left[-\pi y \sin\left(\frac{\pi}{2}(x^2 + y^2)\right), \pi x \sin\left(\frac{\pi}{2}(x^2 + y^2)\right)\right]' $.

 {The discrete numerical scheme is implemented with a spatial grid size of $h = 0.03$ and a time step of $\tau = 0.01$. The temporal evolution is governed by:}
\begin{equation*}
	\begin{cases}
		\frac{\bm{u}^{n+1} - \bm{u}^n}{\tau} + (\bm{u}^n \cdot \nabla)\bm{u}^n - \nu\Delta\bm{u}^{n+1} + \frac{\nabla p^{n+1}}{\rho} = 0, \\
		\frac{p^{n+1} - p^n}{\tau} + \nabla \cdot \bm{u}^{n+1} = 0,
	\end{cases}
\end{equation*}

\begin{figure}[t]
	\centering
	\includegraphics[width=1\linewidth]{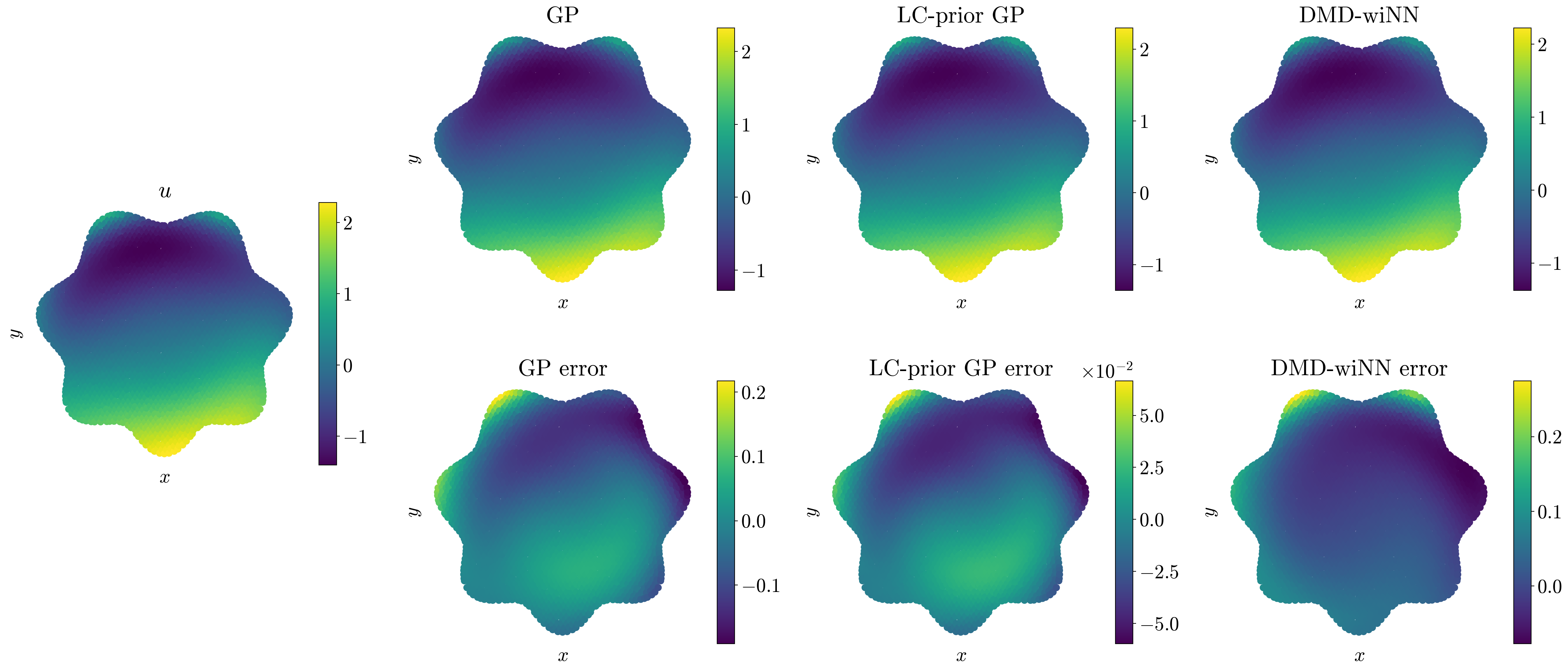}
	\caption{Results for the Navier-Stokes model: mean predictions of x-direction velocity $u(x,y,t;\btheta)$ at $t=0.1$ by different methods and the corresponding pointwise errors.}
	\label{fig:ns_u}
\end{figure}

Here $\btheta= (\mu,\rho)$. The training data is evenly dispersed by 3 points in intervals $\pi_{\text{prior}}(\mu) \sim \mathcal{U}[0,1]$ and $\pi_{\text{prior}}(\rho) \sim \mathcal{U}[0.1,1]$, the physical law corrected data is uniformly dispersed by 5 points and the test data is randomly dispersed by 20 points in the intervals. Thus, the size of the training data is 9, the corrected data contains 16 parameters (excluding the overlapping 9 points of training data) and the test data contains 400 parameters.  

\begin{figure}[t]
	\centering
	\includegraphics[width=1\linewidth]{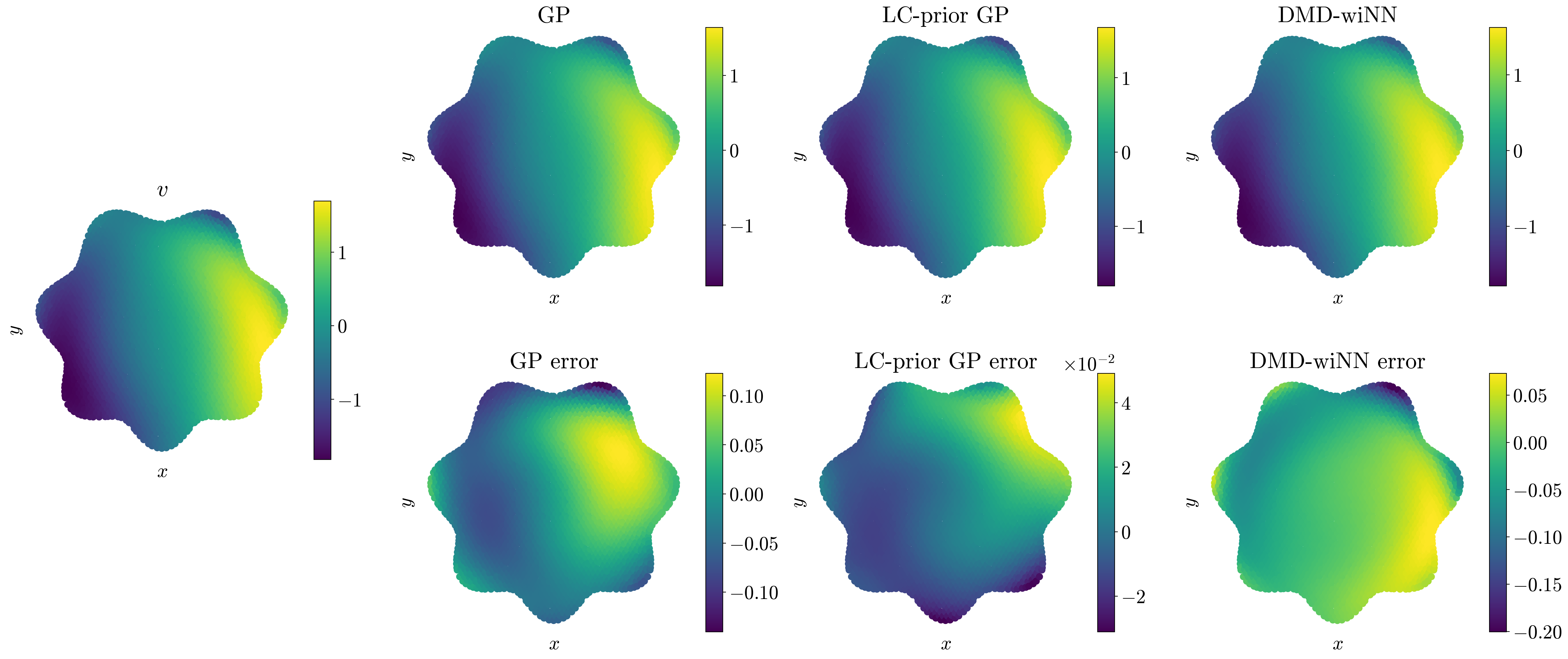}
	\caption{Results for the Navier-Stokes model: mean predictions of y-direction velocity $v(x,y,t;\btheta)$ at $t=0.1$ by different methods and the corresponding pointwise errors.}
	\label{fig:ns_v}
\end{figure}

\begin{figure}[t]
	\centering
	\includegraphics[width=1\linewidth]{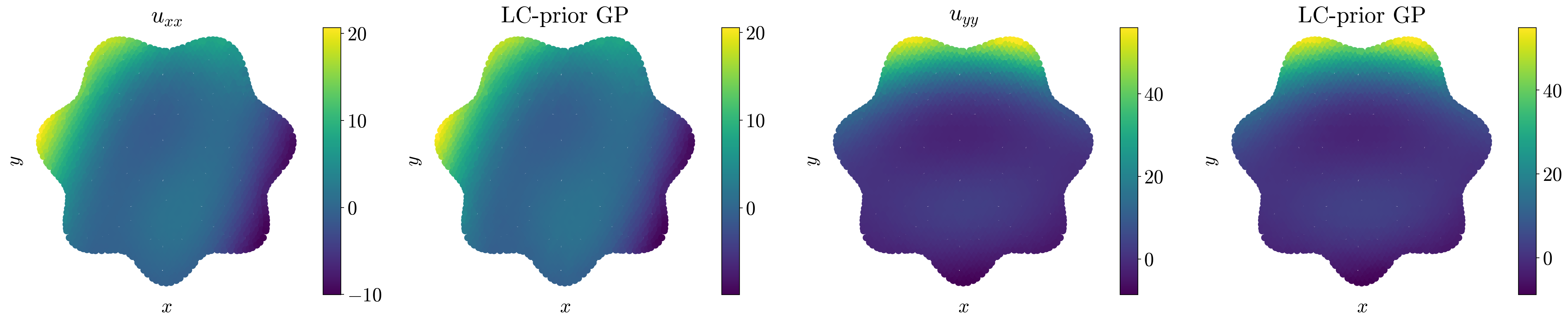}
	\caption{Results for the Navier-Stokes model: the second derivatives of x-direction velocity $u(x,y,t;\btheta)$ at $t=0.1$ obtained by the differentiation matrices.}
	\label{fig:ns_uxx_uyy}
\end{figure}

\begin{figure}[t!]
	\centering
	\includegraphics[width=1\linewidth]{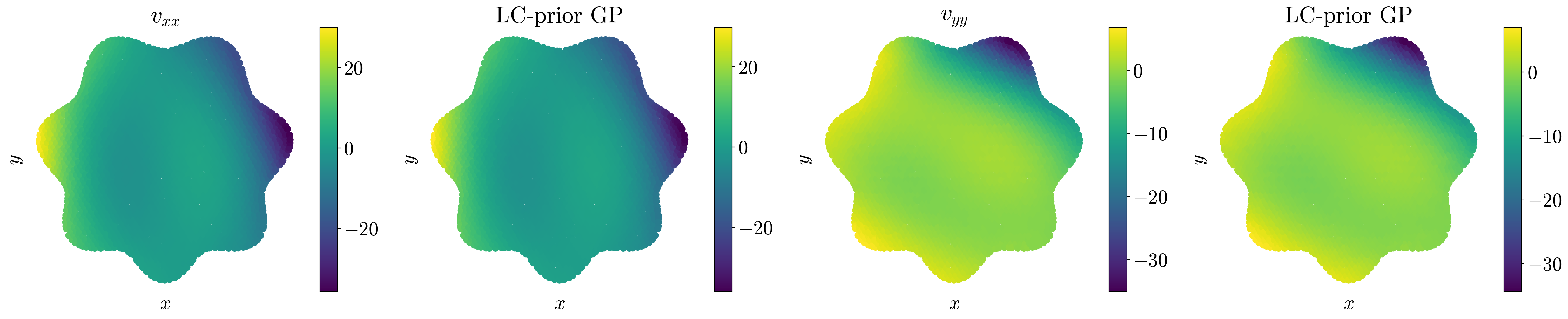}
	\caption{Results for the Navier-Stokes model: the second derivatives of y-direction velocity $v(x,y,t;\btheta)$ at $t=0.1$ obtained by the differentiation matrices.}
	\label{fig:ns_vxx_vyy}
\end{figure}

In this section, we simultaneously construct surrogate models for all three solution fields: the velocity components $u(x,y,t;\btheta)$, $v(x,y,t;\btheta)$, and the pressure field $p(x,y,t;\btheta)$.  {For each physical quantity, the solutions are parameterized using the leading $K=4$ POD modes.} Figure~\ref{fig:ns_u} and Figure~\ref{fig:ns_v} present the computational results for the x-direction and y-direction at $t=0.1$, respectively. Table~\ref{tab:ns_error} provides a comprehensive error analysis, quantifying the aggregate relative errors of the surrogate model across all temporal discretization points. Our proposed method maintains exceptionally small error magnitudes for both velocity components, demonstrating its strong generalization capability for multi-coupled problems.  

Moreover, the precomputed differentiation matrices $D_{h}^{\mathcal{L}_{x}}(X,Y)$ and $D_{h}^{\mathcal{L}_{xx}}(X,Y)$, obtained during training data generation, enable efficient and accurate computation of higher-order derivatives for the target functions. Let $\hat{u}$ and $\hat{v}$ denote the surrogate model predictions, then the second-order derivatives can be calculated by:
$$
\frac{\partial^2 \hat{u}}{\partial x^2} = D_{h}^{\mathcal{L}_{xx}}(X,Y) E^{\dagger}_{h}(Y,X) \hat{u}, \quad
\frac{\partial^2 \hat{v}}{\partial x^2} = D_{h}^{\mathcal{L}_{xx}}(X,Y) E^{\dagger}_{h}(Y,X) \hat{v},
$$
and similarly for $\partial^2 \hat{u}/\partial y^2$ and $\partial^2 \hat{v}/\partial y^2$. Figure~\ref{fig:ns_uxx_uyy} and Figure~\ref{fig:ns_vxx_vyy} present the second-order derivative results along different spatial directions. These results demonstrate that the proposed method maintains efficiency and accuracy in approximating derivatives of the target functions using only matrix operations, providing a natural advantage for learning physics constraints.

\section{Conclusion}
\label{sec:conclusion}
 {In this work, we propose a Gaussian process–based physics-informed framework, termed the physical law-corrected prior GP (LC-prior GP), for surrogate modeling of parametric PDEs, including multi-coupled systems and problems defined on irregular geometries. Proper orthogonal decomposition (POD) is introduced to construct a reduced representation of high-dimensional discrete solutions, ensuring that the GP surrogate is learned in a low-dimensional modal coefficient space. Meanwhile, the governing PDE information is incorporated to learn a more reasonable prior function, thereby correcting the data-driven GP surrogate. By embedding physical constraints into the prior, the proposed method avoids the limitation of conventional physics-informed GP approaches that rely on kernel-based linear operators and are thus restricted to linear PDEs.  
In addition, the RBF-FD method is employed for generating training data, enabling flexible handling of irregular geometries in two-dimensional spaces. Its differentiation matrices can be precomputed and reused in the physics-based correction stage, avoiding repeated evaluations and leading to more efficient optimization compared to other physics-informed machine learning methods. Extensive numerical experiments are conducted for validation, covering multi-parameter, multi-physics variables systems, and different geometric configurations. Comparisons were made with the standard GP, the baseline method and PI-DeepONet (parametric setting) to highlight the efficiency and accuracy.}

\section*{Acknowledgment}
We thank Qiuqi Li and Yuming Ba for their valuable discussions. Heng Yong acknowledges support from the National Science Academic Fund (NSAF) under Grant No. U2230208 and the National Natural Science Foundation of China (NSFC) under Grant No. 12331010. Hongqiao Wang acknowledges support from the NSFC under Grant Nos. 12271562 and 12571470. This work was also supported by the Major Scientific and Technological Innovation Platform Project of Hunan Province (Grant No. 2024JC1003) and was carried out in part using the computing resources of the High Performance Computing Center at Central South University.

\appendix
\section{Non-zero prior mean function for LC-prior GP} \label{sec:non_zero_prior}
 {In common assumptions, we set the prior mean function $m_{k}(\cdot) \equiv 0$, implying no prior knowledge of the outputs. However, in special settings, $m_{k}(\cdot)$ can also be set as the empirical mean of the training data or a parameterized trend function, which is jointly optimized with the kernel hyperparameters. In this section, we focus on the construction of the LC-prior GP when $m_{k}(\cdot) \neq 0$.

Following the setting in the main text, we learn the mapping from parameters to modal coefficients using $K$ independent Gaussian process regressions
$$ 
f_{k}(\btheta) \sim \mathcal{GP}\left(m_{k}(\btheta), \mathbf{k}_{k}(\btheta, \btheta') \right), \quad k=1,\dots,K.
$$
And maximize the log-likelihood function for optimization the hyperparameters:
$$
\log p(\bm{\alpha}_{k}|\btheta, \bm{\zeta}_{k}) = -\frac{1}{2} {(\bm{\alpha}_{k} - \mathbf{m}_k)^\top} \mathbf{K}_{k}^{-1} {(\bm{\alpha}_{k} - \mathbf{m}_k)} - \frac{1}{2} \log \det \mathbf{K}_{k} - \frac{N}{2} \log 2\pi,
$$
where $\mathbf{m}_k = (m_k(\btheta_1), \dots, m_k(\btheta_N))^\top$. It should be noted that when $m_{k}(\cdot)$ is a parameterized trend function optimized at this stage, the likelihood maximization tends to make $\mathbf{m}_k$ closely approximate $\bm{\alpha}_{k}$. As a result, the posterior (predictive) mean function can differ significantly from Eq.~\eqref{eq:gp_poster}. For a new input $\btheta^*$, we have
\begin{equation} \label{eq:non-zero_posterior}
\mu_{k}(\btheta^*) = \mathbb{E}[f_{k}(\btheta^*) \mid \btheta^*, \mathcal{D}_{\text{Low}}] 
= m_{k}(\btheta^*) + \mathbf{k}_{k*}^\top \mathbf{K}^{-1}_{k} \big( \bm{\alpha}_{k} - \mathbf{m}_{k} \big).
\end{equation}
In this case, $(\bm{\alpha}_{k} - \mathbf{m}_{k})$ tends to vanish and the posterior mean becomes highly dependent on the prediction of $m_{k}(\btheta^*)$. If the trend in the test data is inconsistent with that in the training data, the GPR model may be severely misled by the misspecified prior mean $m_{k}(\btheta^*)$. 

Under this setting, the surrogate model $\mathcal{M}_{\text{GP}}$ reads
\begin{equation}
\hat{\bm{u}}(\x;\btheta^*) = \mathcal{M}_{\text{GP}}(\x;\btheta^*) = \sum_{k=1}^{K} \mu_{k}(\btheta^*) \phi_{k}(\x).
\end{equation}
For physical corrected stage, we introduce the corrected function $\omega_{k}(\btheta|\text{Law})$ and physical corrected-prior as
\begin{equation*}
\tilde{m}_k(\btheta|\text{Law}) =
\begin{cases}
	m_{k}(\btheta), & \btheta \in \btheta_{\text{obs}}, \\
	m_{k}(\btheta) + \omega_{k}(\btheta|\text{Law}), & \btheta \in \Theta \setminus \btheta_{\text{obs}}.
\end{cases}
\end{equation*}
And define the LC-prior GP surrogate $\tilde{f}_{k}(\cdot)$ by
$$ 
\tilde{f}_{k}(\btheta) \sim \mathcal{GP}\left(\tilde{m}_{k}(\btheta), \mathbf{k}_{k}(\btheta, \btheta') \right), \quad k=1,\dots,K.
$$
For given any new parameter $\btheta^*$, the posterior mean for LC-prior GP can be rewritten by 
\begin{equation*} 
\begin{aligned}
	   \tilde{\mu}_{k}(\btheta^*)  &= \mathbb{E}[f_{k}(\btheta^*)|\btheta^*, \mathcal{D}_{\text{Low}}, \text{Law}] 
	= \tilde{m}_{k}(\btheta^*|\text{Law}) + \text{\bf{k}}_{k*}^\top \text{\bf{K}}^{-1}_{k} \big( \bm{\alpha}_{k}- \mathbf{m}_{k}  \big)  \\ 
    & = \omega_{k}(\btheta^*|\text{Law}) + m_{k}(\btheta^*) + \text{\bf{k}}_{k*}^\top \text{\bf{K}}^{-1}_{k} \big( \bm{\alpha}_{k}- \mathbf{m}_{k}  \big) 
     = \omega_{k}(\btheta^*|\text{Law}) + {\mu}_{k}(\btheta^*),
\end{aligned}
\end{equation*}
with the reconstruction of target solutions by surrogate model $\mathcal{M}_{\text{LC}}$
\begin{equation*} 
	\begin{aligned}
		\hat{\bm{u}}(\x;\btheta^*) &= \mathcal{M}_{\text{LC}}(\x;\btheta^*) 
		= \sum_{k=1}^{K} \tilde{\mu}_{k}(\btheta^*) \phi_{k}(\x)  
		% = \sum_{k=1}^{K}  \big(\mu_{k}(\btheta^*)+\omega_{k}(\btheta^*) \big) \phi_{k}(\x) 
		= \sum_{k=1}^{K} \omega_{k}(\btheta^*|\text{Law}) \phi_{k}(\x) + \mathcal{M}_{\text{GP}}(\x;\btheta^*).
	\end{aligned}
\end{equation*}
This part is consistent with Eq.~\eqref{eq:lc_posterior} and Eq.~\eqref{eq:M_lc_old}. Therefore, the subsequent optimization based on the loss function Eq.~\eqref{eq:Loss_function} and the learning of the correction mapping $s_{k}(\cdot)$ can be carried out following the same procedure as shown in Algorithm~\ref{Algorithm:lc-prior_gp}, and will not be repeated here.

This result indicates that the LC-prior GP remains self-consistent even when a non-zero prior mean function is adopted, with derivations identical to the case where \(m_{k}(\cdot) \equiv 0\). The main point to note is that, when the prior includes a parameterized trend function jointly optimized with the kernel hyperparameters, the maximum log-likelihood tends to drive \((\bm{\alpha}_{k} - \mathbf{m}_{k})\) toward zero. Consequently, the posterior mean in Eq.~\eqref{eq:non-zero_posterior} becomes largely dominated by the prediction of \(m_{k}(\btheta^*)\). In practice, however, full consistency between the input-output distributions of training and test data is rarely guaranteed. Therefore, assuming \(m_{k}(\cdot) \equiv 0\) and focusing on kernel optimization along with the physical correction is generally a more reasonable and conservative choice.}

\begingroup
\small 
\bibliographystyle{unsrt}
\bibliography{ref}

@article{kondo2010reaction,
  title={Reaction-diffusion model as a framework for understanding biological pattern formation},
  author={Kondo, S. and Miura, T.},
  journal={Science},
  volume={329},
  number={5999},
  pages={1616--1620},
  year={2010}
}

@article{ewing2001summary,
  title={A summary of numerical methods for time-dependent advection-dominated partial differential equations},
  author={Ewing, R. E. and Wang, H.},
  journal={J. Comput. Appl. Math.},
  volume={128},
  number={1-2},
  pages={423--445},
  year={2001}
}

@book{boyer2012mathematical,
  title={Mathematical Tools for the Study of the Incompressible Navier-Stokes Equations and Related Models},
  author={Boyer, F. and Fabrie, P.},
  volume={183},
  year={2012},
  publisher={Springer}
}

@article{andrieu2008tutorial,
  title={A tutorial on adaptive {MCMC}},
  author={Andrieu, C. and Thoms, J.},
  journal={Stat. Comput.},
  volume={18},
  number={4},
  pages={343--373},
  year={2008}
}

@article{chib1995understanding,
  title={Understanding the {M}etropolis-{H}astings algorithm},
  author={Chib, S. and Greenberg, E.},
  journal={Am. Stat.},
  volume={49},
  number={4},
  pages={327--335},
  year={1995}
}

@article{zhu1997algorithm,
  title={Algorithm 778: L-BFGS-B: Fortran subroutines for large-scale bound-constrained optimization},
  author={Zhu, C. and Byrd, R. H. and Lu, P. and Nocedal, J.},
  journal={ACM Trans. Math. Softw.},
  volume={23},
  number={4},
  pages={550--560},
  year={1997}
}

@book{quarteroni2015reduced,
  title={Reduced basis methods for partial differential equations: an introduction},
  author={Quarteroni, A. and Manzoni, Andrea and Negri, Federico},
  year={2015},
  publisher={Springer}
}

@article{pfortner2022physics,
  title={Physics-informed {G}aussian process regression generalizes linear {PDE} solvers},
  author={Pf{\"o}rtner, M. and Steinwart, I. and Hennig, P. and Wenger, J.},
  journal={arXiv preprint arXiv:2212.12474},
  year={2022}
}

@article{persson2004simple,
  title={A simple mesh generator in {MATLAB}},
  author={Persson, P.-O. and Strang, G.},
  journal={SIAM Rev.},
  volume={46},
  number={2},
  pages={329--345},
  year={2004}
}

@book{tarantola2005inverse,
  title={Inverse problem theory and methods for model parameter estimation},
  author={Tarantola, A.},
  year={2005},
  publisher={SIAM}
}

@article{lu2021learning,
  title={Learning nonlinear operators via {D}eep{ON}et based on the universal approximation theorem of operators},
  author={Lu, L. and Jin, P. and Pang, G. and Zhang, Z. and Karniadakis, G. E.},
  journal={Nat. Mach. Intell.},
  volume={3},
  number={3},
  pages={218--229},
  year={2021}
}

@book{williams2006gaussian,
  title={Gaussian processes for machine learning},
  author={Williams, C. K. I. and Rasmussen, C. E.},
  year={2006},
  publisher={MIT Press}
}

@article{karniadakis2021physics,
  title={Physics-informed machine learning},
  author={Karniadakis, G. E. and Kevrekidis, I. G. and Lu, L. and Perdikaris, P. and Wang, S. and Yang, L.},
  journal={Nat. Rev. Phys.},
  volume={3},
  number={6},
  pages={422--440},
  year={2021}
}

@article{wang2021learning,
  title={Learning the solution operator of parametric partial differential equations with physics-informed {D}eep{ON}ets},
  author={Wang, S. and Wang, H. and Perdikaris, P.},
  journal={Sci. Adv.},
  volume={7},
  number={40},
  pages={eabi8605},
  year={2021}
}

@article{chen2021physics,
  title={Physics-informed machine learning for reduced-order modeling of nonlinear problems},
  author={Chen, W. and Wang, Q. and Hesthaven, J. S. and Zhang, C.},
  journal={J. Comput. Phys.},
  volume={446},
  pages={110666},
  year={2021}
}

@article{lu2022comprehensive,
  title={A comprehensive and fair comparison of two neural operators (with practical extensions) based on fair data},
  author={Lu, L. and Meng, X. and Cai, S. and Mao, Z. and Goswami, S. and Zhang, Z. and Karniadakis, G. E.},
  journal={Comput. Methods Appl. Mech. Eng.},
  volume={393},
  pages={114778},
  year={2022}
}

@incollection{pang2020physics,
  title={Physics-informed learning machines for partial differential equations: Gaussian processes versus neural networks},
  author={Pang, G. and Karniadakis, G. E.},
  booktitle={Emerging frontiers in nonlinear science},
  pages={323--343},
  year={2020},
  publisher={Springer}
}

@article{wang2021explicit,
  title={Explicit estimation of derivatives from data and differential equations by {G}aussian process regression},
  author={Wang, H. and Zhou, X.},
  journal={Int. J. Uncertain. Quantif.},
  volume={11},
  number={4},
  year={2021}
}

@book{dhatt2012finite,
  title={Finite element method},
  author={Dhatt, G. and Lefranc{c}ois, E. and Touzot, G.},
  year={2012},
  publisher={John Wiley \& Sons}
}

@book{thomas2013numerical,
  title={Numerical partial differential equations: finite difference methods},
  author={Thomas, J. W.},
  volume={22},
  year={2013},
  publisher={Springer}
}

@article{raissi2019physics,
  title={Physics-informed neural networks: A deep learning framework for solving forward and inverse problems involving nonlinear partial differential equations},
  author={Raissi, M. and Perdikaris, P. and Karniadakis, G. E.},
  journal={J. Comput. Phys.},
  volume={378},
  pages={686--707},
  year={2019}
}

@article{sirignano2018dgm,
  title={DGM: A deep learning algorithm for solving partial differential equations},
  author={Sirignano, J. and Spiliopoulos, K.},
  journal={J. Comput. Phys.},
  volume={375},
  pages={1339--1364},
  year={2018}
}

@article{chen2022bridging,
  title={Bridging traditional and machine learning-based algorithms for solving {PDE}s: the random feature method},
  author={Chen, J. and Chi, X. and Yang, Z. and others},
  journal={J. Mach. Learn.},
  volume={1},
  number={3},
  pages={268--298},
  year={2022}
}

@book{temam2024navier,
  title={Navier--Stokes equations: theory and numerical analysis},
  author={Temam, R.},
  volume={343},
  year={2024},
  publisher={Amer. Math. Soc.}
}

@article{yu2022gradient,
  title={Gradient-enhanced physics-informed neural networks for forward and inverse {PDE} problems},
  author={Yu, J. and Lu, L. and Meng, X. and Karniadakis, G. E.},
  journal={Comput. Methods Appl. Mech. Eng.},
  volume={393},
  pages={114823},
  year={2022}
}

@article{kennedy2001bayesian,
  title={Bayesian calibration of computer models},
  author={Kennedy, M. C. and O'Hagan, A.},
  journal={J. R. Stat. Soc. Ser. B (Stat. Methodol.)},
  volume={63},
  number={3},
  pages={425--464},
  year={2001}
}

@article{mora2025operator,
  title={Operator learning with {G}aussian processes},
  author={Mora, C. and Yousefpour, A. and Hosseinmardi, S. and Owhadi, H. and Bostanabad, R.},
  journal={Comput. Methods Appl. Mech. Eng.},
  volume={434},
  pages={117581},
  year={2025}
}

@article{shankar2017overlapped,
  title={The overlapped radial basis function-finite difference (RBF-FD) method: A generalization of {RBF-FD}},
  author={Shankar, V.},
  journal={J. Comput. Phys.},
  volume={342},
  pages={211--228},
  year={2017}
}

@article{bayona2010rbf,
  title={RBF-FD formulas and convergence properties},
  author={Bayona, V. and Moscoso, M. and Carretero, M. and Kindelan, M.},
  journal={J. Comput. Phys.},
  volume={229},
  number={22},
  pages={8281--8295},
  year={2010}
}

@article{rudy2019data,
  title={Data-driven identification of parametric partial differential equations},
  author={Rudy, S. and Alla, A. and Brunton, S. L. and Kutz, J. N.},
  journal={SIAM J. Appl. Dyn. Syst.},
  volume={18},
  number={2},
  pages={643--660},
  year={2019}
}

@article{tripura2023wavelet,
  title={Wavelet neural operator for solving parametric partial differential equations in computational mechanics problems},
  author={Tripura, T. and Chakraborty, S.},
  journal={Comput. Methods Appl. Mech. Eng.},
  volume={404},
  pages={115783},
  year={2023}
}

@article{song2024model,
  title={A model reduction method for parametric dynamical systems defined on complex geometries},
  author={Song, H. and Ba, Y. and Chen, D. and Li, Q.},
  journal={J. Comput. Phys.},
  volume={506},
  pages={112923},
  year={2024}
}

@book{schiesser2019time,
  title={Time Delay ODE/PDE Models: Applications in Biomedical Science and Engineering},
  author={Schiesser, W. E.},
  year={2019},
  publisher={CRC Press}
}

@article{pestourie2023physics,
  title={Physics-enhanced deep surrogates for partial differential equations},
  author={Pestourie, R. and Mroueh, Y. and Rackauckas, C. and Das, P. and Johnson, S. G.},
  journal={Nat. Mach. Intell.},
  volume={5},
  number={12},
  pages={1458--1465},
  year={2023}
}

@article{schaback2024using,
  title={Using compactly supported radial basis functions to solve partial differential equations},
  author={Schaback, R. and Wendland, H.},
  journal={WIT Trans. Model. Simul.},
  volume={23},
  year={2024}
}

@article{nguyen2023proper,
  title={Proper orthogonal descriptors for efficient and accurate interatomic potentials},
  author={Nguyen, N. C. and Rohskopf, A.},
  journal={J. Comput. Phys.},
  volume={480},
  pages={112030},
  year={2023}
}

@article{nekkanti2023gappy,
  title={Gappy spectral proper orthogonal decomposition},
  author={Nekkanti, A. and Schmidt, O. T.},
  journal={J. Comput. Phys.},
  volume={478},
  pages={111950},
  year={2023}
}

@article{pichi2024graph,
  title={A graph convolutional autoencoder approach to model order reduction for parametrized {PDE}s},
  author={Pichi, F. and Moya, B. and Hesthaven, J. S.},
  journal={J. Comput. Phys.},
  volume={501},
  pages={112762},
  year={2024}
}

@article{chen2021improved,
  title={An improved data-free surrogate model for solving partial differential equations using deep neural networks},
  author={Chen, X. and Chen, R. and Wan, Q. and Xu, R. and Liu, J.},
  journal={Sci. Rep.},
  volume={11},
  number={1},
  pages={19507},
  year={2021}
}

@article{radaideh2020surrogate,
  title={Surrogate modeling of advanced computer simulations using deep {G}aussian processes},
  author={Radaideh, M. I. and Kozlowski, T.},
  journal={Reliab. Eng. Syst. Saf.},
  volume={195},
  pages={106731},
  year={2020}
}

@article{mishra2018machine,
  title={A machine learning framework for data driven acceleration of computations of differential equations},
  author={Mishra, S.},
  journal={arXiv preprint arXiv:1807.09519},
  year={2018}
}

@article{lucia2004reduced,
  title={Reduced-order modeling: new approaches for computational physics},
  author={Lucia, D. J. and Beran, P. S. and Silva, W. A.},
  journal={Prog. Aerosp. Sci.},
  volume={40},
  number={1-2},
  pages={51--117},
  year={2004}
}

@article{de2013basis,
  title={A basis for bounding the errors of proper generalised decomposition solutions in solid mechanics},
  author={de Almeida, J. P. M.},
  journal={Int. J. Numer. Methods Eng.},
  volume={94},
  number={10},
  pages={961--984},
  year={2013}
}

@article{baur2011interpolatory,
  title={Interpolatory projection methods for parameterized model reduction},
  author={Baur, U. and Beattie, C. and Benner, P. and Gugercin, S.},
  journal={SIAM J. Sci. Comput.},
  volume={33},
  number={5},
  pages={2489--2518},
  year={2011}
}

@article{hesthaven2018non,
  title={Non-intrusive reduced order modeling of nonlinear problems using neural networks},
  author={Hesthaven, J. S. and Ubbiali, S.},
  journal={J. Comput. Phys.},
  volume={363},
  pages={55--78},
  year={2018}
}

@article{berkooz1993proper,
  title={The proper orthogonal decomposition in the analysis of turbulent flows},
  author={Berkooz, G. and Holmes, P. and Lumley, J. L.},
  journal={Annu. Rev. Fluid Mech.},
  volume={25},
  number={1},
  pages={539--575},
  year={1993}
}

@article{brivio2024ptpi,
  title={PTPI-DL-ROMs: Pre-trained physics-informed deep learning-based reduced order models for nonlinear parametrized PDEs},
  author={Brivio, S. and Fresca, S. and Manzoni, A.},
  journal={Comput. Methods Appl. Mech. Eng.},
  volume={432},
  pages={117404},
  year={2024}
}
\endgroup
\end{document}